\documentclass[preprint,12pt]{elsarticle-arxiv}
\pdfoutput=1

\usepackage{tikz}
\usepackage{cite}
\usepackage{ifpdf}
\usepackage{multirow}
\usepackage{afterpage}
\usepackage{booktabs}
\usepackage{amssymb}
\usepackage{amsthm}
\usepackage{amsmath}
\usepackage{pdflscape}
\usepackage[vlined,ruled,commentsnumbered]{algorithm2e}
\usepackage{subcaption}

\makeatletter
\newcommand\footnoteref[1]{\protected@xdef\@thefnmark{\ref{#1}}\@footnotemark}
\makeatother

\usepackage{url}

\newcommand{\posset}{\mathcal{P}}
\newcommand{\unlset}{\mathcal{U}}

\newcommand{\resvm}{RESVM}

\journal{Neurocomputing: SI on Advances in Learning with Label Noise\hfill 20/10/2014}
\begin{document}

\begin{frontmatter}
 \title{A Robust Ensemble Approach to Learn From Positive and Unlabeled Data Using SVM Base Models}
 
 \author[stadius]{Marc Claesen\corref{cor1}}
 \ead{marc.claesen@esat.kuleuven.be}
 
 \author[pubhealth]{Frank De Smet\fnref{fn1}}
 \ead{frank.desmet@cm.be}
 
 \author[stadius]{Johan A. K. Suykens}
 \ead{johan.suykens@esat.kuleuven.be}

 \author[stadius]{Bart De Moor}
 \ead{bart.demoor@esat.kuleuven.be} 
 
 \cortext[cor1]{Corresponding author. Tel.: +32 16 328649.}
 \fntext[fn1]{Frank De Smet is a member of the medical management department of the 
National Alliance of Christian Mutualities.}
 
 \address[stadius]{KU Leuven, ESAT -- STADIUS/iMinds Medical IT  \\
Kasteelpark Arenberg 10, box 2446 \\
3001 Leuven, Belgium}
 \address[pubhealth]{KU Leuven, Department of Public Health and Primary Care, Environment and Health \\
Kapucijnenvoer 35 blok d, box 7001 \\
3000 Leuven, Belgium}

\begin{abstract}%
We present a novel approach to learn binary classifiers when only positive and unlabeled instances are available (PU learning). This problem is routinely cast as a supervised task with label noise in the negative set. We use an ensemble of SVM models trained on bootstrap resamples of the training data for increased robustness against label noise. The approach can be considered in a bagging framework which provides an intuitive explanation for its mechanics in a semi-supervised setting. We compared our method to state-of-the-art approaches in simulations using multiple public benchmark data sets. The included benchmark comprises three settings with increasing label noise: (i) fully supervised, (ii) PU learning and (iii) PU learning with false positives. Our approach shows a marginal improvement over existing methods in the second setting and a significant improvement in the third.
\end{abstract}


\end{frontmatter}


\section{Introduction}

Training binary classifiers on positive and unlabeled data is referred to as PU learning \citep{Liu:2003:BTC:951949.952139}. The absence of known negative training instances warrants appropriate learning methods. Inaccurate label information can be more problematic than attribute noise \citep{zhu2004class}. Specialised PU learning approaches are recommended when (i) negative labels cannot be acquired, (ii) the training data contains a large amount of false negatives or (iii) the positive set has many outliers.

Practical applications of PU learning typically feature large, imbalanced training sets with a small amount of labeled (positive) and a large amount of unlabeled training instances. The PU learning problem arises in various settings, including web page classification \citep{Yu02pebl:positive}, intrusion detection \citep{Lazarevic03acomparative} and bioinformatics tasks such as variant prioritization \citep{sifrim2013extasy}, gene prioritization \citep{citeulike:615632, mordelet2011prodige} and virtual screening of drug compounds \citep{citeulike:3911}. 

Though these applications share a common underlying learning problem, the final evaluation criteria may be fundamentally different. For instance, in  prioritization one wishes to obtain high precision since highly ranked targets may be subjected to further biological analysis. Intrusion detection, on the other hand, necessitates high recall to ensure that no anomalies go unnoticed.

Following \citet{MORDELET-2010-523336}, we will use the term \emph{contamination} to refer to the fraction of mislabeled instances in a given set. We will denote the positive and unlabeled training instances by $\mathcal{P}$ and $\mathcal{U}$, respectively. Contamination in $\mathcal{P}$ refers to false positives while contamination in $\mathcal{U}$ refers to the presence of positives in $\mathcal{U}$. Usually $\mathcal{U}$ contains mostly true negative instances (e.g. contamination below $0.5$) and $\mathcal{P}$ is assumed to be uncontaminated. 

The distributions of the positive and a contaminated unlabeled set overlap even when those of the positive and underlying negative sets do not, which makes classification more difficult compared to a traditional supervised setting. \citet{Elkan:2008:LCO:1401890.1401920} and \citet{blanchard2010semi} report statistical approaches to estimate the contamination of the unlabeled set and additionally show that distinguishing positives from unlabeled instances is a valid proxy for distinguishing positives from negatives.

The assumption in PU learning that $\mathcal{P}$ is uncontaminated may be violated in applications due to various reasons \citep{frenay}. Additionally, outliers in the positive set may have a similar effect on classification performance \citep{pechenizkiy2006class}. We propose a novel PU learning method that is less vulnerable to potential contamination in $\mathcal{P}$ called the robust ensemble of support vector machines (RESVM). RESVM is compared to other methods in a series of simulations based on several public data sets.


\section{Related work}
PU learning approaches can be split into two main conceptual categories: (i) approaches that account for the contamination of the unlabeled set explicitly by modeling the label noise and (ii) approaches that try to infer an uncontaminated (negative) subset $\hat{\mathcal{N}}$ from $\unlset$ and then train supervised algorithms to distinguish $\posset$ from $\hat{\mathcal{N}}$. When \emph{very} few labeled examples are available, the structure within the data is the main source of information which can be exploited by semi-supervised clustering techniques \citep{bksc2}. 

\paragraph{Accounting for the contamination of $\mathcal{U}$ in the modeling process}
This can be done by weighting individual data points, such as in weighted logistic regression \citep{Elkan:2008:LCO:1401890.1401920,Lee03learningwith}. Another approach is by changing the penalties on misclassification during training, as is done in class-weighted SVM \citep{Liu:2003:BTC:951949.952139}, bagging SVM \citep{MORDELET-2010-523336} and RT-SVM \citep{Liu:2005:PSC:2138033.2138052}. 

\paragraph{Inferring an uncontaminated subset from $\mathcal{U}$}
Another class of approaches tries to infer a negative set $\hat{\mathcal{N}}$ from $\mathcal{U}$. After the inferential step, binary classifiers are trained to distinguish $\mathcal{P}$ from $\mathcal{\hat{N}}$ in a supervised fashion. Examples of such two-step approaches include S-EM \citep{liu02partially}, mapping convergence (MC) \citep{Yu:2005:SCM:1108759.1108762} and ROC-SVM \citep{Li03learningto}.

\paragraph{Class-weighted SVM and related approaches} The approach we suggest belongs to the first class of methods and is closely related to class-weighted SVM and bagging SVM (which uses class-weighted SVM internally). We will discuss both of these approaches in more detail before moving on to the proposed method. We evaluated our method compared to both class-weighted SVM and bagging SVM.

\subsection{Class-weighted SVM} \label{bsvm}
Class-weighted SVM (CWSVM) is a supervised technique in which the penalty for misclassification differs per class. \citet{Liu:2003:BTC:951949.952139} first applied class-weighted SVM for PU learning by considering the unlabeled set to be negative with noise on its labels. CWSVM is trained to distinguish $\mathcal{P}$ from $\mathcal{U}$. During training, misclassification of positive instances is penalized more than misclassification of unlabeled instances to emphasize the higher degree of certainty on positive labels. In the context of PU learning, the optimization problem for training CWSVM can be written as:
\begin{align}
\min_{\alpha,\xi,b}\ & \frac{1}{2}\sum_{i=1}^N\sum_{j=1}^N \alpha_i\alpha_j y_i y_j \kappa(\mathbf{x}_i,\mathbf{x}_j)+C_{\mathcal{P}}\sum_{i \in\mathcal{P}} \xi_i + C_{\mathcal{U}}\sum_{i\in\mathcal{U}} \xi_i, \label{eq:bsvm} \\
\text{s.t. } &y_i(\sum_{j=1}^N \alpha_j y_j \kappa(\mathbf{x}_i,\mathbf{x}_j)+b)\geq 1-\xi_i, &i=1,\ldots,N, \nonumber \\
&\xi_i \geq 0, &i=1,\ldots,N, \nonumber
\end{align}
with $\alpha \in \mathbb{R}^N$ the support values, $\mathbf{y} \in \{-1,+1\}^N$ the label vector, $\kappa(\cdot,\cdot)$ the kernel function, $b$ the bias term and $\xi \in \mathbb{R}^N$ the slack variables. The misclassification penalties $C_{\mathcal{P}}$ and $C_{\mathcal{U}}$ require tuning (typically $C_{\mathcal{P}} > C_{\mathcal{U}}$ to emphasize known labels). SVM formulations with unequal penalties across classes have been used previously to tackle imbalanced data sets \citep{osuna1997}. 

\subsection{Bagging SVM} \label{baggingsvm}
Mordelet and Vert introduce bagging SVM as a meta-algorithm which consists of aggregating classifiers trained to discriminate $\mathcal{P}$ from a small, random resample of\ $\mathcal{U}$ \citep{MORDELET-2010-523336}. They posit that PU learning problems have a particular structure that leads to instability of classifiers, namely the sensitivity of classifiers to the contamination of the unlabeled set.  Bagging is a common technique used to improve the performance of instable classifiers \citep{Breiman:1996:BP:231986.231989}.

In bagging SVM, random resamples of $\unlset$ are drawn and CWSVM classifiers are trained to discriminate $\posset$ from each resample. By resampling $\unlset$, the contamination is varied. This induces variability in the classifiers which the aggregation procedure can then exploit. The size of the bootstrap resample of $\mathcal{U}$ is a tuning parameter in bagging SVM. The ratio $C_{\mathcal{P}}/C_{\mathcal{U}}$ is fixed so that the following holds:
\begin{equation}
|\mathcal{P}| \times C_{\mathcal{P}} = n_{\mathcal{U}} \times C_{\mathcal{U}}, \label{eq:bagpenalties}
\end{equation}
with $|\mathcal{P}|$ the size of the positive set and $n_{\mathcal{U}}$ the size of resamples from the unlabeled set. This choice of weights is common in imbalanced settings \citep{cawley2006leave,daemen2009kernel}. All base models in bagging SVM classify the full set of positives against a subset of unlabeled instances and use a high misclassification penalty on the positives similar to CWSVM. 


\section{Robust Ensemble of SVMs}

We propose a new technique called the robust ensemble of SVMs (RESVM). RESVM is a bagging method using CWSVM base models as discussed in Section~\ref{bsvm}. Base model training sets are constructed by bootstrap resampling both $\mathcal{P}$ and $\mathcal{U}$ separately, both of which may be contaminated. 

The key difference between RESVM and bagging SVM is that the former resamples $\mathcal{P}$ in addition to $\mathcal{U}$ to increase variability between base models. RESVM additionally features an extra degree of freedom to control the relative misclassification penalty between positive and unlabeled instances, which is fixed in bagging SVM. \citet{MORDELET-2010-523336} report no significant changes when varying the relative penalty in bagging SVM, though our experiments show that it is important in RESVM (see $w_{pos}$ in Table~\ref{table:hyperparameters}).

Before elaborating on the details of RESVM, we briefly illustrate the effect of resampling contaminated sets. Subsequently we summarize the mechanisms of bagging and why they are advantageous when learning with label noise in the RESVM approach. Finally, we provide the full RESVM training approach and the way ensemble decision values are computed based on the base model decision values.


\subsection{Bootstrap resampling contaminated sets} \label{resampling}

\begin{figure}[!h]
  \centering
  \includegraphics[width=0.80\textwidth]{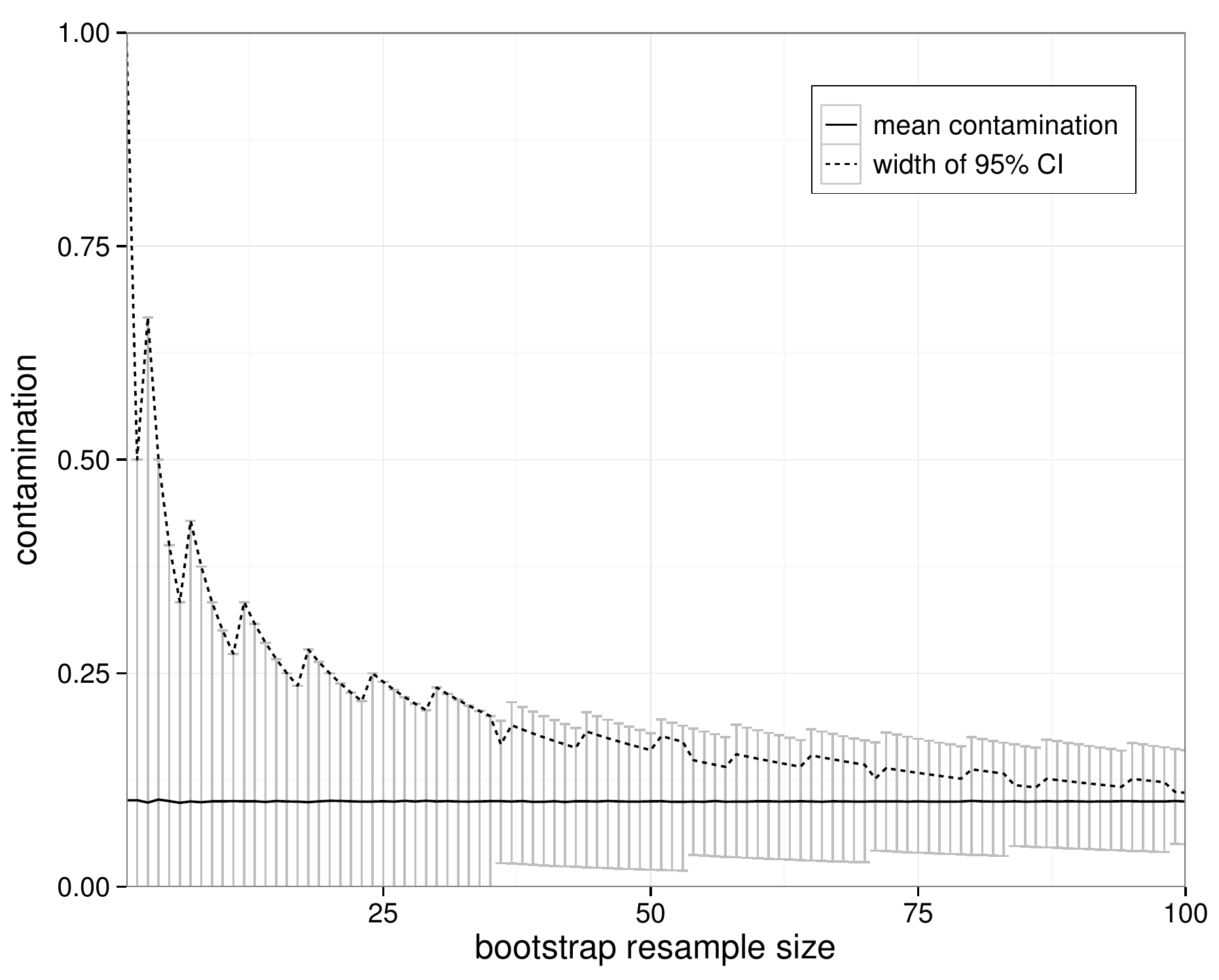}
  \caption{Contamination of bootstrap resamples for increasing size of resamples when the original sample has $10\%$ contamination. Errorbars indicate the $95\%$ confidence interval (CI) of contamination in resamples. The contamination varies greatly between small resamples as shown by the CIs. } 
  \label{fig:contamination}
\end{figure}

The RESVM approach resamples both $\mathcal{P}$ and $\mathcal{U}$, both of which are potentially contaminated. Resampling contaminated sets with replacement induces variability in contamination across the resampled sets (e.g. resamples of $\mathcal{U}$ and $\mathcal{P}$ that are used for training). The variability in contamination between resamples increases for increasing contamination of the original set. We assume contamination levels below $50\%$, e.g. less than half the instances in a given set are mislabeled. Due to the law of large numbers the contamination in bootstrap resamples of increasing size converges to the expected contamination, which equals that of the original set that is being resampled. As a result, the variability in contamination decreases for increasing resample size. Figure~\ref{fig:contamination} illustrates this property empirically based on $20,000$ repeated measurements for each resample size: the expected value (mean) equals the original contamination, but the variability in resample contamination decreases for increasing resample size.

\subsection{Bagging predictors}
\citet{Breiman:1996:BP:231986.231989} introduced bagging as a technique to construct strong ensembles by combining a set of base models. \citet{breiman2000randomizing} stated that ``the essential problem in combining classifiers is growing a suitably diverse ensemble of base classifiers'' which can be done in various ways \citep{brown2005diversity}. In bagging, the ensemble models use majority voting to aggregate decisions of base models which are trained on bootstrap resamples of the training set. From a Bayesian point of view, bagging can be interpreted as a Monte Carlo integration over an approximated posterior distribution \citep{rao1997out}. 

In his landmark paper, \citet{Breiman:1996:BP:231986.231989} noted that base model instability is an important factor in the success of bagging which led to the use of inherently instable methods like decision trees in early bagging approaches \citep{dietterich2000experimental, breiman2001random}. The main mechanism of bagging is often said to be variance reduction \citep{bauer1999empirical,breiman2000randomizing}. In more recent work, \citet{grandvalet2004bagging} explained that base model instability is not related to the intrinsic variability of a predictor but rather to the presence of influential instances in a data set for a given predictor (so-called \emph{leverage points}). The effect of bagging is explained as equalizing the influence of all training instances, which is beneficial when highly influential instances are harmful for the predictor's accuracy. 


\subsection{Justification of the RESVM algorithm}
We have shown the effect of resampling contaminated sets and provided some basic insight into the mechanics of bagging. We will now link these two elements to justify bagging approaches in the context of contaminated training sets. Its usefulness can be considered by both the variance reduction argument of \citet{bauer1999empirical} and equalizing the influence of training points as described by \citet{grandvalet2004bagging}.

\paragraph{Variance reduction} Resampling a contaminated set yields different levels of contamination in the resamples as explained in Section~\ref{resampling}. Varying the contamination between base model training sets induces variability between base models without increasing bias. This observation enables us to create a diverse set of base models by resampling both $\mathcal{P}$ and $\mathcal{U}$. The variance reduction of bagging is an excellent mechanism to exploit the variability of base models based on resampling \citep{bauer1999empirical,breiman2000randomizing}. In the context of RESVM, a tradeoff takes place between increased variability (by training on smaller resamples, see Figure~\ref{fig:contamination}) and base models with increased stability (larger training sets for the SVM models).

\paragraph{Equalizing influence} The influence of a training instance on an SVM model can be quantified in terms of its dual weight (the associated $\alpha$ value). Three distinct cases can be distinguished: (i) the training instance is correctly classified and not within the margin ($\alpha = 0$, not a SV), (ii) the training instance lies on the margin and is correctly classified ($\alpha \in [0, C]$, free SV) and (iii) the training instance is incorrectly classified or within the margin ($\alpha = C$, bounded SV), where $C$ is the misclassification penalty associated to the training instance \citep{bottou2007support}. Instances that are misclassified during training become bounded SVs, which have the maximal $\alpha$ value and can therefore be considered leverage points of the SVM model. When learning with label noise, the mislabeled training instances are likely to end up as bounded SVs. In a best case scenario, the mislabeled training instances are classified in concordance to their true label by the SVM model (which means they must be a bounded SV as the training procedure identifies this as a misclassification). As such, mislabeled training instances act as leverage points for SVM models. Following \citet{grandvalet2004bagging}, bagging equalizes the influence of training instances (e.g. lowers the influence of mislabeled leverage points in comparison to the rest of the data) which yields improved robustness against contamination in the context of RESVM. 


\subsection{RESVM training}
RESVM uses CWSVM base models trained on resamples from the original training set, where both $\mathcal{P}$ and $\mathcal{U}$ are being resampled. The technique involves $5$ hyperparameters: $3$ to define the resampling strategy and $2$ for the base models. Additional hyperparameters may be involved, for example $\gamma$ for the RBF kernel $\kappa(\mathbf{x}_i, \mathbf{x}_j) = \exp(-\gamma \| \mathbf{x}_i-\mathbf{x}_j\|^2)$.

The number of base models to include in the ensemble, $n_{nmodels}$, is the first hyperparameter. Using more base models improves the stability of the ensemble (up to a certain plateau) at a linear increase in computational cost for training and prediction. $n_{models}$ is not a traditional hyperparameter in the sense that a good value can be determined during training, for example based on out-of-bag error estimates \citep{banfield2007comparison}.\footnote{\label{labelednote}Note that the error estimates in out-of-bag techniques must account for potential contamination. See our discussion of hyperparameter tuning for a possible score function.}

By resampling $\mathcal{P}$, RESVM takes potential contamination of the labeled instances into account by design. Since the contamination between $\mathcal{P}$ and $\mathcal{U}$ can vary, the ability to vary the size of resamples from $\mathcal{P}$ and $\mathcal{U}$ separately is required. This results in two tuning parameters: $n_{pos}$ and $n_{unl}$. In general, using small base model training sets results in increased base model variability which then necessitates using more base models in the ensemble to obtain a given level of stability. In our experiments, we have tuned $n_{pos}$ and $n_{unl}$ but it is also possible to obtain good values using out-of-bag techniques \citep{martinez2010out}.\footnoteref{labelednote}

RESVM additionally inherits at least $2$ hyperparameters from its SVM base models, namely misclassification penalties for both classes and, if applicable, hyperparameters related to the kernel function. We define the CWSVM penalties in see Eq.~\eqref{eq:bsvm} based on $2$ hyperparameters $C_{\mathcal{U}}$ and $w_{pos}$:
\begin{equation}
C_{\mathcal{P}} = C_{\mathcal{U}} \times w_{pos} \times \frac{n_{unl}}{n_{pos}}. \label{eq:resvmpenalties}
\end{equation}

$w_{pos}$ enables reweighting labeled and unlabeled instances after equalizing class imbalance. In bagging SVM, $w_{pos}$ is always fixed to $1$. 

\newpage
The RESVM training approach has been summarised in Algorithm~\ref{RESVM}. The algorithm uses $5$ hyperparameters plus additional kernel parameters.

\newlength\mydatalen
\newcommand\mydata[1]{%
  \settowidth\mydatalen{\KwData{}}%
  \setlength\hangindent{\mydatalen}%
  \hspace*{\mydatalen}#1\\}

\newlength\myinputlen
\newcommand\myinput[1]{%
  \settowidth\myinputlen{\KwIn{}}%
  \setlength\hangindent{\myinputlen}%
  \hspace*{\myinputlen}#1\\}

\begin{algorithm}[!h]
\DontPrintSemicolon
\KwData{$\mathcal{P}$: the set of positive instances.}
\mydata{$\mathcal{U}$: the set of unlabeled instances.}
\KwIn{$n_{models}$: number of base models to include in the ensemble.}
\myinput{$n_{unl}$: size of bootstrap resamples of $\mathcal{U}$.}
\myinput{$n_{pos}$: size of bootstrap resamples of $\mathcal{P}$.}
\myinput{$C_{\mathcal{U}}$: misclassification penalty for $\mathcal{U}$ in class-weighted SVM.}
\myinput{$w_{pos}$: relative positive misclassification penalty coefficient.}
\myinput{$\kappa(\cdot,\cdot)$: kernel function to be used by base models.}
\KwOut{$\Omega$: RESVM with $n_{models}$ base models.} 
\Begin{
$\Omega \leftarrow \emptyset$; \;
$C_{\mathcal{P}} \leftarrow C_{\mathcal{U}} \times w_{pos} \times \frac{n_{unl}}{n_{pos}}$; \;
\For{$i \leftarrow 1$ \KwTo $n_{models}$}{
\texttt{// create base model training set from $\mathcal{P}$ and $\mathcal{U}$.} \;
$\mathcal{P}^{(i)} \leftarrow$ sample $n_{pos}$ instances from $\mathcal{P}$ with replacement;\;
$\mathcal{U}^{(i)} \leftarrow$ sample $n_{unl}$ instances from $\mathcal{U}$ with replacement;\;
\texttt{// train CWSVM base model $\psi^{(i)}$ and add to ensemble $\Omega$.} \;
$\psi^{(i)} \leftarrow$ train CWSVM for $\mathcal{P}^{(i)}$ vs. $\mathcal{U}^{(i)}$ (parameters $C_{\mathcal{P}}$, $C_{\mathcal{U}}$, $\kappa$); \;
$\Omega \leftarrow \{\Omega,\psi^{(i)}\}$; \; 
}
}
\caption{Training procedure for RESVM\label{RESVM}.} \label{RESVM}
\end{algorithm}


\subsection{RESVM prediction} \label{ensembledecvals}
RESVM uses majority voting to aggregate base model predictions. By default, the returned label is the one predicted by most base models. The fraction of positive votes for a test instance $\mathbf{x}$ can be written as:
\begin{equation}
v(\mathbf{x}) = \frac{n_{models} + \sum_{i=1}^{n_{models}} \mathtt{sgn}(\psi^{(i)}(\mathbf{x}))}{2 n_{models}},
\label{eq:majorityvote}
\end{equation}
where $\mathtt{sgn}(\cdot)$ is the sign function and $\psi^{(i)}$ denotes the decision function of SVM base model $i$ with codomain $\mathbb{R}$. $v(\cdot)$ has the interval $[0,1]$ as codomain.

The RESVM decision value for a test instance $\mathbf{x}$ is defined as the fraction of votes in favor of the positive class $v(\mathbf{x})$ unless the result is unanimous. In the case of a unanimous vote, the ensemble decision value is based on the decision values of its base models to increase the model's ability to differentiate. In case of a unanimous negative vote, the sum of the decision values of the base models is taken (each SVM base model decision value is negative in this case). In case of a unanimous positive vote, the sum of the decision values of the base models (all positive) plus one is taken.
The decision value $d(\cdot)$ has codomain $\mathbb{R}$ and is computed as follows:
\begin{align}
d(\mathbf{x}) &= \left\{ \begin{array}{ll}
v(\mathbf{x}) & \text{if } 0 < v(\mathbf{x}) < 1, \\
\sum_{i=1}^{n_{models}} \psi^{(i)}(\mathbf{x}) & \text{if } v(\mathbf{x}) = 0, \\
1+\sum_{i=1}^{n_{models}} \psi^{(i)}(\mathbf{x}) & \text{if } v(\mathbf{x}) = 1.
\end{array}\right. \label{eq:resvmdecval}
\end{align}
The resulting label for a given decision threshold $T$ can be written as follows:
\begin{equation}
l(\mathbf{x}) = \mathtt{sgn}\big(d(\mathbf{x})-T\big). \label{eq:resvmlabel}
\end{equation}
The default decision value threshold for positive classification is $T=0.5$ (this is majority voting, e.g. positive iff more than half of all base models predict positive). Using the modified decision values $d(\mathbf{x})$ instead of the votes $v(\mathbf{x})$ does not affect the predicted labels for typical choices of the threshold $T$ (e.g. $T \in (0, 1)$). It does, however, affect performance measures that use the entire range of decision values such as area under the PR curve. Using $d(\mathbf{x})$ enables us to rank different instances that received all positive or all negative votes by base models (e.g. $v(\mathbf{x})=1$ and $v(\mathbf{x})=0$, respectively).


\section{Experimental setup}
\resvm\ has been compared to class-weighted SVM (CWSVM) and bagging SVM (BAG) in a number of simulations to assess the merits of our modifications compared to conceptually comparable algorithms. In this Section we will summarize the experimental setup (training set construction, model selection and performance evaluation) and the data sets we used. 

\subsection{Simulation setup} 
\begin{figure}[!h]
  \centering
  \includegraphics[width=\textwidth]{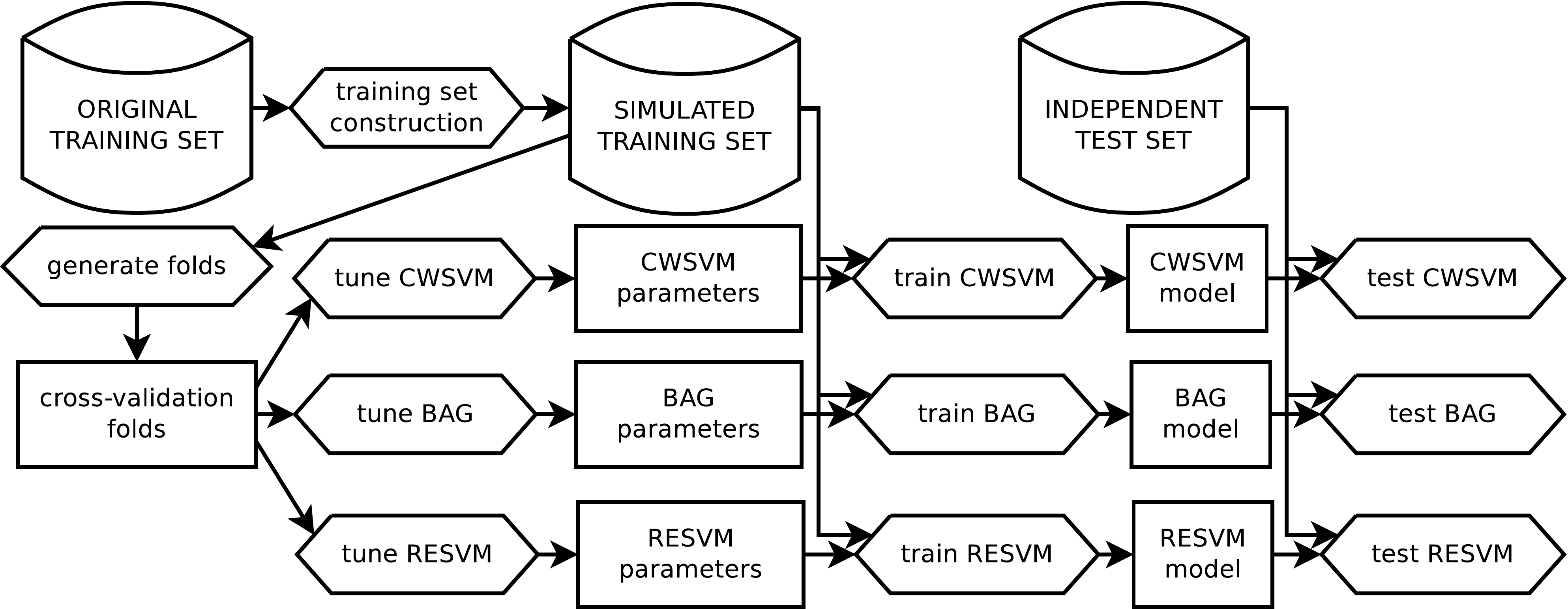}
  \caption{Overview of a single benchmark iteration.} 
  \label{fig:benchmark}
\end{figure}

Our experiments consist of repeated simulations on a variety of data sets under different settings. Briefly, in each iteration hyperparameters were optimized per approach based on cross-validation on the training set (using identical folds for all approaches). Subsequently, a model with the optimal parameters is trained on the full training set and used to predict an independent test set. An overview of the experiments is shown in Figure~\ref{fig:benchmark}. Every experiment consists of 20 repetitions. 

To assess what situations are favorable per approach, we have investigated three different settings with distinct label noise configurations. For every data set, we performed 10 iterations per simulation in the following settings:
\begin{enumerate}
\item \textbf{supervised}: no contamination in $\mathcal{P}$ or $\mathcal{U}$ ($\mathcal{U}$ is the negative class).
\item \textbf{PU learning}: contamination in $\mathcal{U}$ but not in $\mathcal{P}$. 
\item \textbf{semi-supervised}: contamination in both $\mathcal{P}$ and $\mathcal{U}$. The contamination levels in $\mathcal{P}$ and $\mathcal{U}$ were always chosen equal.
\end{enumerate}

The contamination levels we used were chosen per data set based on when differences between the three approaches become visible. A summary is available in Table~\ref{table:datasets} in Section~\ref{data}. When applicable, contamination was introduced by flipping class labels (e.g. true positives in $\mathcal{U}$ and true negatives in $\mathcal{P}$). This effectively changes the empirical densities of the classes in the training set (illustrated in Figure~\ref{fig:densities} in the next Section).

Every binary learning task was repeated 20 times to get reliable assessments of all methods. Each repetition involves redoing all steps shown in Figure~\ref{fig:benchmark}, including resampling of training sets based on the known true positives and true negatives. Contamination was introduced at random where applicable by flipping class labels.


\paragraph{Hyperparameter selection}
In every iteration, hyperparameters were tuned per setting using 10-fold cross-validation over a grid of parameter tuples. To ensure a fair comparison, one set of folds is generated in each iteration and used by all methods. We ensured that the optimal values that were found during tuning in any setting were never on the edge of the search grid. The search resolution in comparable parameters between methods was always defined to be identical (for example $\gamma$ in the case of an RBF kernel). 

The same search grids were used in all three settings for a given data set to illustrate that a method can work well in a supervised setting with a given search grid but degrade when label noise is added. Since negative labels are unavailable in PU learning, we used the following score function in all learning settings which only requires positive labels for hyperparameter selection \citep{Lee03learningwith}:
\begin{equation}
{\tt pu\_score} = \frac{\mathtt{precision}\times \mathtt{recall}}{Pr(y=1)} =
\frac{\mathtt{recall}^2}{Pr(\hat{y}=1)},
\label{puscore}
\end{equation}
where $Pr(y=1)$ is the fraction of known positive labels in the predicted set and $Pr(\hat{y}=1)$ is the fraction of positive predictions made by the classifier. Note that this score function is not ideal when $\mathcal{P}$ is contaminated, though we obtained good results even in that setting.

The following parameters were tuned per method: (CWSVM) $C_{\mathcal{P}}$ and $C_{\mathcal{U}}$, (BAG) $C_{\mathcal{U}}$ and $n_{\mathcal{U}}$ and (RESVM) $C_{\mathcal{U}}$, $w_{pos}$, $n_{pos}$ and $n_{unl}$. In both ensemble approaches we consistently used 50 base models.


\paragraph{Performance assessment} Models are trained with the optimal hyperparameters on the full training set and subsequently tested on the independent test set. We use the known test labels to compute the area under the Precision-Recall curve (AUC) for each model. We opted to use PR curves because they capture the performance of interest of models over their entire operating range and work well for imbalanced data \citep{Davis:2006:RPR:1143844.1143874}. 

We used statistical analyses to determine whether one approach trumps another while accounting for the variability between simulations. 
The nonparametric Wilcoxon signed-rank test is recommended for pairwise comparisons between learning algorithms \citep{demvsar2006statistical}. In every setting per data set we performed a paired one-tailed Wilcoxon signed-rank test comparing the area under the PR curve of bagging SVM and RESVM with alternative hypothesis $h_1: AUC^{\resvm} > AUC^{BAG}$ (pairs being iterations). Low $p$-values indicate a statistically significant improvement. 


\paragraph{Implementation details}
We used the class-weighted SVM implementation available in LIBLINEAR \citep{fan2008liblinear} and LIBSVM \citep{CC01a} for models using the linear and RBF kernel, respectively. Bagging SVM and \resvm\ were implemented using the EnsembleSVM library \citep{claesen14}.\footnote{Python code for RESVM is available at \url{https://github.com/claesenm/resvm}.} The decision values of bagging SVM used to compute PR curves were defined in the same way as for RESVM (see Section~\ref{ensembledecvals}).


\subsection{Data sets} \label{data} 
\begin{figure}[!h]
  \centering
  \includegraphics[width=0.40\textwidth]{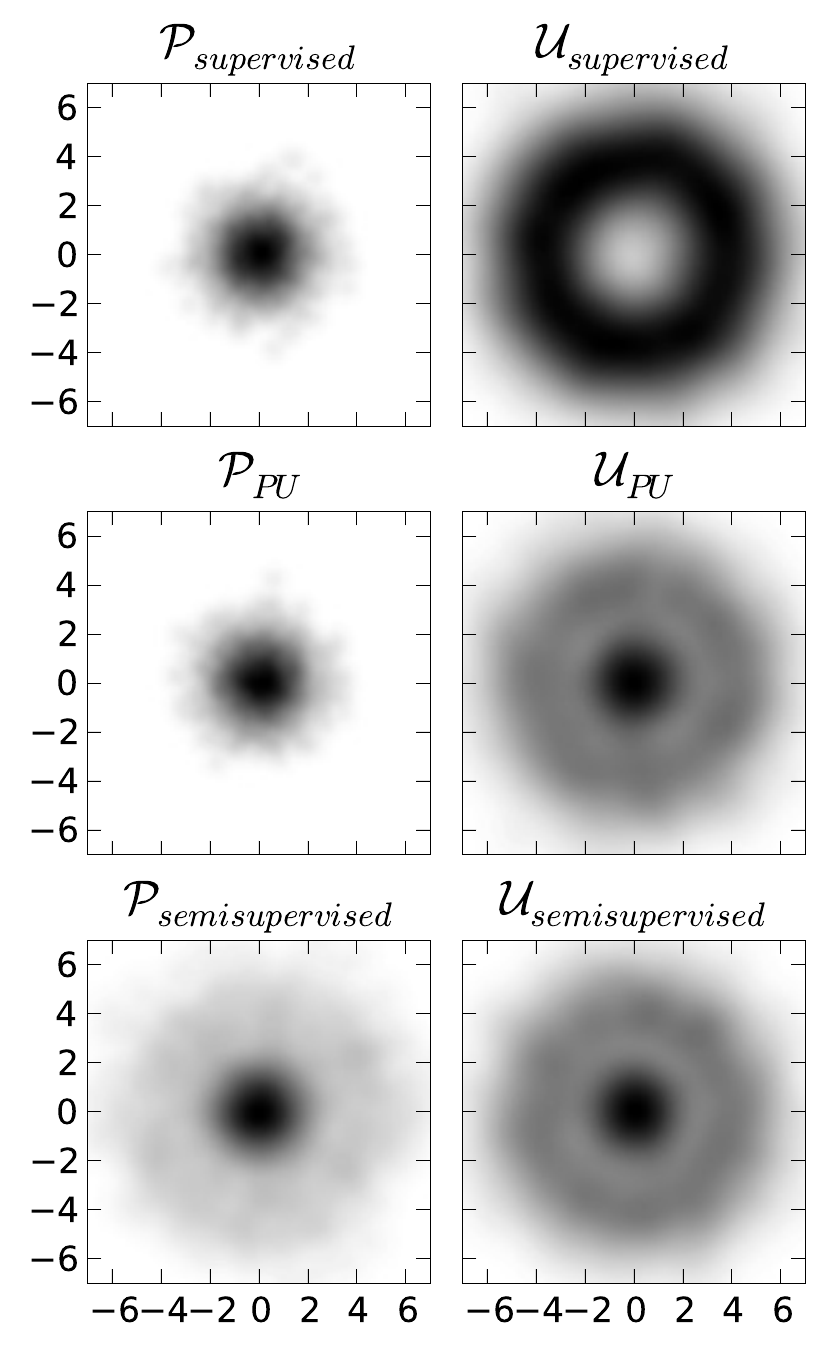}
  \caption{Empirical densities of the \texttt{synthetic} data used for training per problem setting (visualized in input space). The supervised densities (top row) are based on samples of the underlying positive and negative classes. The use of high contamination ($30\%$) induces similar empirical densities for $\mathcal{P}$ and $\mathcal{U}$ in the semi-supervised setting (bottom row).} 
  \label{fig:densities}
\end{figure}

We used a synthetic data set and 5 publicly available data sets:\footnote{Public data at: \url{http://www.csie.ntu.edu.tw/~cjlin/libsvmtools/datasets/}.}
\begin{itemize}
\item \texttt{synthetic}: a 2-D binary data set. Positive instances are sampled from a standard normal distribution. Negative instances are sampled from a circle centered at the origin with radius 4 with 2-D noise superimposed from a standard normal distribution. Training and testing data was generated in every iteration. Figure~\ref{fig:densities} shows densities for all settings. 
\item \texttt{cancer}: the Wisconsin breast cancer data set related to breast cancer diagnosis. It consists of $10$ features and $683$ instances without an explicit train/test partitioning so we partitioned it at random in every iteration.
\item \texttt{ijcnn1}: used for the IJCNN 2001 neural network competition \citep{prokhorov2001ijcnn}, comprising $2$ classes, $22$ features and $49,990/91,701$ training/testing instances.
\item \texttt{covtype}: a common classification benchmark about predicting forest cover types based on cartographic information \citep{Blackard00covtype}. We used a subsample of $100,000/40,000$ training/testing instances.
\item \texttt{mnist}: a digit recognition task \citep{Lecun98gradient-basedlearning}. This data set contains 10 classes (one for each digit), $780$ features, $60,000$ training instances and $10,000$ test instances with an almost uniform class distribution. We performed one-versus-all classification for each digit.
\item \texttt{sensit}: SensIT Vehicle (combined), vehicle classification \citep{duarte2004vehicle}. This data set contains 3 classes with an uneven distribution. We performed one-versus-all classification for each class. This data set has $100$ features, $78,823$ training instances and $19,705$ testing instances.
\end{itemize}

Most data sets have a prespecified test set, except for \texttt{synthetic} and \texttt{cancer}. We used the prespecified test sets when available. We used the RBF kernel for all data sets except \texttt{mnist} (linear kernel). Note that both RESVM and bagging SVM models are always implicitly nonlinear due to their majority voting scheme, even when using linear base models.

In every setting each original data set was resampled without replacement to construct training sets to use in the simulations. The resampled training sets are typically significantly smaller than what is available in the original data sets to show that some methods can obtain good models even with few training instances. An overview of the actual training sets we constructed is presented in Table~\ref{table:datasets}.

\begin{table}[!ht]
\centering
\begin{tabular}{crcrrcrr}
\toprule
 & & contamination & \multicolumn{2}{c}{training set} & & \multicolumn{2}{c}{test set} \\ \cline{4-5} \cline{7-8}
data set & $d$ & in percent & $|\mathcal{P}|$ & $|\mathcal{U}|$ & & $|\mathcal{P}|$ & $|\mathcal{N}|$ \\
\midrule
\texttt{synthetic} & $2$ & $30$ & $100$ & $200$ & & $5,000$ & $5,000$ \\
\texttt{cancer} & $10$ & $30$   & $50$ & $200$ & & $100$ & $100$ \\
\texttt{ijcnn1} & $22$ & $10$   & $100$ & $10,000$ & & $8,712$ & $82,989$ \\
\texttt{covtype} & 54 & 30 & 100 & 1,000 & & 20,000 & 20,000 \\
\texttt{mnist} & $780$ & $10$   & $50$ & $2,000$ & & $\approx1,000$ & $\approx9,000$ \\
\texttt{sensit 1} & $100$ & $30$  & $100$ & $1,000$ & & $4,575$ & $15,130$ \\
\texttt{sensit 2} & $100$ & $30$  & $100$ & $1,000$ & & $5,520$ & $14,455$ \\
\texttt{sensit 3} & $100$ & $30$  & $100$ & $1,000$ & & $9,880$ & $9,825$ \\
\bottomrule
\end{tabular}
\caption{Overview of the data sets used in simulations: number of features, contamination (when applicable), training set size as used in the experiments and test set size. The \texttt{mnist} data set consists of 10 classes and the test set is almost uniformly distributed. The \texttt{sensit} data set has 3 classes with uneven class distribution in the test set, so we treat it separately here.}

\label{table:datasets}
\end{table}


\newpage
\section{Results and discussion}
We will summarize all results of our simulation experiments comparing class-weighted SVM (CWSVM), bagging SVM (BAG) and the robust ensemble of SVMs (RESVM). First we will show the results of each setting separately. Subsequently we present an overview of the number of wins per setting for each method across all data sets. Section~\ref{varycontamination} shows the results of an experiment to assess the effect of contamination in $\mathcal{P}$ and $\mathcal{U}$ on all methods. Finally, we include an interesting observation regarding the optimal hyperparameters of RESVM that were found using cross-validation on the \texttt{mnist} data set per setting in Table~\ref{table:hyperparameters}.
\subsection{Results for supervised classification}
Table~\ref{table:supervised} summarizes our results in a fully supervised setting. In these experiments both $\mathcal{P}$ and $\mathcal{U}$ are uncontaminated. Based on the number of wins per simulation and the confidence intervals, we can conclude that all methods are competitive in this setting. 

The confidence intervals show that all methods obtain comparable results for all simulations except \texttt{mnist} digit $8$, where CWSVM performs poorly compared to the others. This performance difference could be caused by the fact we used linear class-weighted SVM while both ensemble methods implicitly yield nonlinear decision boundaries. A linear model may be too simple to properly distinguish this digit from the others.

The overall good results in the supervised setting confirm that the score function in Equation~\eqref{puscore} is a good choice for tuning. In these supervised experiments we could have used a traditional score like accuracy, area under the ROC curve or F-measure, but these would no longer be useful in the other settings. The performance in these supervised experiments can be considered an objective baseline for comparison in the PU learning and semi-supervised setting since only levels of contamination are varied.

\begin{table}[!h]
\centering
\begin{tabular}{cccccccc}
\toprule
 & \multicolumn{3}{c}{area under PR curve} & & \multicolumn{3}{c}{number of wins} \\ \cline{2-4} \cline{6-8}
data & \texttt{CWSVM} & \texttt{BAG} & \texttt{RESVM} & $p$ & \texttt{CWSVM} & \texttt{BAG} & \texttt{RESVM} \\
\midrule
\texttt{synthetic} & $98.1$--$98.7$ & $98.7$--$98.8$ & $98.7$--$98.8$ &  & 2 & 12 & 6\\
\texttt{cancer} & $98.4$--$98.8$ & $98.4$--$98.7$ & $98.3$--$98.7$ &  & 8 & 12 & 0\\
\texttt{ijcnn1} & $85.3$--$87.4$ & $79.1$--$81.6$ & $82.3$--$86.2$ & $\bullet\ \bullet\ \bullet$ & 16 & 0 & 4\\
\texttt{covtype} & $77.1$--$78.3$ & $76.8$--$78.5$ & $76.8$--$78.7$ &  & 8 & 6 & 6\\ 
\multicolumn{3}{l}{\texttt{mnist} (positive = \texttt{x})} \\
\texttt{0} & $96.9$--$97.5$ & $96.9$--$97.4$ & $96.9$--$97.4$ &  & 7 & 8 & 5\\ 
\texttt{1} & $98.1$--$98.3$ & $98.3$--$98.5$ & $98.2$--$98.5$ &  & 0 & 8 & 12\\ 
\texttt{2} & $87.3$--$89.1$ & $88.5$--$89.8$ & $89.6$--$90.5$ & $\bullet$ & 2 & 6 & 12\\ 
\texttt{3} & $83.7$--$85.9$ & $86.9$--$88.7$ & $88.8$--$90.1$ & $\bullet\ \bullet\ \bullet$ & 0 & 5 & 15\\ 
\texttt{4} & $88.8$--$90.2$ & $89.8$--$91.1$ & $90.8$--$92.2$ & $\bullet\ \bullet\ \bullet$ & 1 & 3 & 16\\ 
\texttt{5} & $78.7$--$80.9$ & $79.2$--$81.0$ & $81.4$--$83.2$ & $\bullet\ \bullet$ & 3 & 3 & 14\\ 
\texttt{6} & $92.4$--$93.4$ & $93.9$--$94.7$ & $94.3$--$94.9$ &  & 0 & 8 & 12\\ 
\texttt{7} & $92.2$--$92.9$ & $92.6$--$93.2$ & $93.1$--$93.7$ & $\bullet\ \bullet\ \bullet$ & 1 & 3 & 16\\ 
\texttt{8} & $56.5$--$58.9$ & $74.3$--$76.1$ & $79.6$--$80.5$ & $\bullet\ \bullet\ \bullet$ & 0 & 0 & 20\\ 
\texttt{9} & $72.5$--$75.6$ & $77.8$--$80.3$ & $81.5$--$82.6$ & $\bullet\ \bullet\ \bullet$ & 0 & 2 & 18\\ 
\multicolumn{3}{l}{\texttt{sensit} (positive = \texttt{x})} \\
\texttt{1} & $80.5$--$81.4$ & $79.8$--$80.7$ & $80.5$--$81.3$ & $\bullet$ & 10 & 2 & 8\\ 
\texttt{2} & $65.7$--$75.4$ & $72.6$--$74.0$ & $73.5$--$74.9$ & $\bullet\ \bullet\ \bullet$ & 15 & 0 & 5\\ 
\texttt{3} & $35.5$--$56.1$ & $92.3$--$92.7$ & $91.7$--$92.3$ &  & 0 & 15 & 5\\ 
\bottomrule
\end{tabular}
\caption{$95\%$ CIs for mean test set performance in a fully supervised setup, the results of a paired one-tailed Wilcoxon signed-rank test comparing the AUC of \texttt{BAG} and \texttt{\resvm} with alternative hypothesis $h_1: AUC^{\resvm} > AUC^{BAG}$ and the number of times each approach had best test set performance.
Test result encoding: $\bullet$ $p < 0.05$, $\bullet\ \bullet$ $p < 0.01$ and $\bullet\ \bullet\ \bullet$ $p < 0.001$.
}
\label{table:supervised}
\end{table}


\subsection{Results for PU learning}
The results of our experiments in a PU learning setting are shown in Table~\ref{table:pulearning}. In the pure PU learning setting, $\mathcal{P}$ is uncontaminated but $\mathcal{U}$ is contaminated. Class-weighted SVM tends to suffer from the largest loss in performance between supervised learning and pure PU learning based on area under PR curves. Class-weighted SVM obtains less wins than it did in the supervised simulations (21 wins in PU learning compared to 73 in the supervised setting), except on the \texttt{cancer} data set. Bagging SVM and RESVM maintain strong performance. Bagging SVM obtains a comparable number of wins and RESVM gains many compared to the supervised setting.

On the \texttt{mnist} data, RESVM consistently exhibits the best performance (based on the Wilcoxon signed-rank test), though the effective improvement over bagging SVM is marginal. On \texttt{sensit} with classes 2 or 3 as positive, bagging SVM obtains the majority of wins though the confidence intervals of its area under the PR curve overlap completely with those of RESVM. On the other data sets, no worthwile differences were obtained between both ensemble methods.

\begin{table}[!h]
\centering
\begin{tabular}{cccccccc}
\toprule
 & \multicolumn{3}{c}{area under PR curve} & & \multicolumn{3}{c}{number of wins} \\ \cline{2-4} \cline{6-8}
data & \texttt{CWSVM} & \texttt{BAG} & \texttt{RESVM} & $p$ & \texttt{CWSVM} & \texttt{BAG} & \texttt{RESVM} \\
\midrule
\texttt{synthetic} & $96.9$--$98.4$ & $97.9$--$98.6$ & $98.2$--$98.5$ &  & 6 & 8 & 6\\ 
\texttt{cancer} & $98.2$--$98.5$ & $87.5$--$98.4$ & $96.1$--$98.1$ &  & 10 & 7 & 3\\
\texttt{ijcnn1} & $71.2$--$76.5$ & $73.4$--$78.2$ & $72.6$--$80.7$ & $\bullet$ & 1 & 5 & 14\\
\texttt{covtype} & $65.2$--$67.9$ & $70.2$--$72.2$ & $71.4$--$73.0$ &  & 0 & 6 & 14\\ 
\multicolumn{3}{l}{\texttt{mnist} (positive = \texttt{x})} \\
\texttt{0} & $74.1$--$77.8$ & $90.5$--$93.3$ & $94.6$--$95.5$ & $\bullet\ \bullet\ \bullet$ & 0 & 5 & 15\\ 
\texttt{1} & $89.1$--$91.2$ & $95.2$--$96.7$ & $96.4$--$97.3$ & $\bullet\ \bullet$ & 0 & 5 & 15\\ 
\texttt{2} & $55.2$--$60.1$ & $75.5$--$80.0$ & $84.2$--$86.1$ & $\bullet\ \bullet\ \bullet$ & 0 & 0 & 20\\ 
\texttt{3} & $54.6$--$60.2$ & $74.5$--$80.3$ & $83.6$--$86.2$ & $\bullet\ \bullet\ \bullet$ & 0 & 2 & 18\\ 
\texttt{4} & $57.8$--$62.5$ & $73.9$--$80.3$ & $83.9$--$85.9$ & $\bullet\ \bullet\ \bullet$ & 0 & 2 & 18\\ 
\texttt{5} & $53.3$--$56.7$ & $63.8$--$70.3$ & $69.1$--$72.6$ & $\bullet$ & 0 & 7 & 13\\ 
\texttt{6} & $66.9$--$71.0$ & $85.9$--$89.7$ & $90.6$--$92.5$ & $\bullet\ \bullet$ & 0 & 4 & 16\\ 
\texttt{7} & $71.4$--$74.8$ & $84.0$--$88.0$ & $90.0$--$91.4$ & $\bullet\ \bullet\ \bullet$ & 0 & 1 & 19\\ 
\texttt{8} & $34.8$--$38.8$ & $63.5$--$69.1$ & $72.2$--$74.8$ & $\bullet\ \bullet\ \bullet$ & 0 & 4 & 16\\ 
\texttt{9} & $50.5$--$54.8$ & $66.2$--$71.0$ & $74.2$--$76.4$ & $\bullet\ \bullet\ \bullet$ & 0 & 1 & 19\\ 
\multicolumn{3}{l}{\texttt{sensit} (positive = \texttt{x})} \\
$\texttt{1}$ & $61.6$--$73.0$ & $70.6$--$75.3$ & $72.5$--$76.2$ & $\bullet$ & 2 & 7 & 11\\ 
$\texttt{2}$ & $58.6$--$68.1$ & $68.5$--$70.5$ & $67.8$--$70.0$ &  & 2 & 10 & 8\\ 
$\texttt{3}$ & $33.2$--$50.2$ & $90.2$--$91.8$ & $89.7$--$91.1$ &  & 0 & 14 & 6\\ 
\bottomrule
\end{tabular}
\caption{$95\%$ CIs for mean test set performance in a PU learning setup, the results of a paired one-tailed Wilcoxon signed-rank test comparing the AUC of \texttt{BAG} and \texttt{\resvm} with alternative hypothesis $h_1: AUC^{\resvm} > AUC^{BAG}$ and the number of times each approach had best test set performance.
Test result encoding: $\bullet$ $p < 0.05$, $\bullet\ \bullet$ $p < 0.01$ and $\bullet\ \bullet\ \bullet$ $p < 0.001$.
}
\label{table:pulearning}
\end{table}

\begin{table}[!h]
\centering
\begin{tabular}{cccccccc}
\toprule
 & \multicolumn{3}{c}{area under PR curve} & & \multicolumn{3}{c}{number of wins} \\ \cline{2-4} \cline{6-8}
data & \texttt{CWSVM} & \texttt{BAG} & \texttt{RESVM} & $p$ & \texttt{CWSVM} & \texttt{BAG} & \texttt{RESVM} \\
\midrule
\texttt{synthetic} & $83.6$--$90.0$ & $91.9$--$94.9$ & $96.4$--$97.4$ & $\bullet\ \bullet\ \bullet$ & 3 & 2 & 15\\
\texttt{cancer} & $62.5$--$80.2$ & $91.1$--$96.7$ & $96.2$--$97.6$ & $\bullet$ & 1 & 8 & 11\\ 
\texttt{ijcnn1} & $69.8$--$73.4$ & $67.4$--$70.4$ & $72.0$--$75.2$ & $\bullet\ \bullet\ \bullet$ & 5 & 2 & 13\\
\texttt{covtype} & $58.1$--$61.8$ & $61.2$--$64.2$ & $60.4$--$65.7$ &  & 4 & 4 & 12\\ 
\multicolumn{3}{l}{\texttt{mnist} (positive = \texttt{x})} \\
\texttt{0} & $59.9$--$64.1$ & $72.8$--$81.1$ & $91.4$--$93.4$ & $\bullet\ \bullet\ \bullet$ & 0 & 0 & 20\\ 
\texttt{1} & $80.3$--$82.7$ & $90.6$--$93.4$ & $96.1$--$97.4$ & $\bullet\ \bullet\ \bullet$ & 0 & 0 & 20\\ 
\texttt{2} & $42.3$--$48.0$ & $55.1$--$63.7$ & $79.8$--$83.0$ & $\bullet\ \bullet\ \bullet$ & 0 & 0 & 20\\ 
\texttt{3} & $43.8$--$47.6$ & $59.9$--$66.0$ & $78.1$--$81.1$ & $\bullet\ \bullet\ \bullet$ & 0 & 0 & 20\\ 
\texttt{4} & $52.4$--$56.2$ & $66.4$--$72.8$ & $79.7$--$83.4$ & $\bullet\ \bullet\ \bullet$ & 0 & 0 & 20\\ 
\texttt{5} & $40.5$--$45.2$ & $56.0$--$61.1$ & $65.8$--$69.4$ & $\bullet\ \bullet\ \bullet$ & 0 & 2 & 18\\ 
\texttt{6} & $52.4$--$57.3$ & $72.9$--$79.3$ & $87.9$--$90.9$ & $\bullet\ \bullet\ \bullet$ & 0 & 0 & 20\\ 
\texttt{7} & $58.7$--$61.6$ & $69.9$--$77.3$ & $87.9$--$90.2$ & $\bullet\ \bullet\ \bullet$ & 0 & 1 & 19\\ 
\texttt{8} & $29.7$--$33.9$ & $48.3$--$55.3$ & $68.0$--$71.0$ & $\bullet\ \bullet\ \bullet$ & 0 & 0 & 20\\ 
\texttt{9} & $42.1$--$44.9$ & $52.5$--$59.0$ & $68.7$--$72.7$ & $\bullet\ \bullet\ \bullet$ & 0 & 0 & 20\\ 
\multicolumn{3}{l}{\texttt{sensit} (positive = \texttt{x})} \\
$\texttt{1}$ & $34.5$--$49.4$ & $59.6$--$69.0$ & $60.6$--$66.4$ &  & 3 & 12 & 5\\ 
$\texttt{2}$ & $44.9$--$53.7$ & $46.4$--$53.4$ & $50.1$--$56.7$ & $\bullet$ & 8 & 4 & 8\\ 
$\texttt{3}$ & $44.5$--$61.1$ & $75.4$--$83.5$ & $80.5$--$84.9$ & $\bullet$ & 1 & 7 & 12\\ 
\bottomrule
\end{tabular}
\caption{$95\%$ CIs for mean test set performance in a semi-supervised setup, the results of a paired one-tailed Wilcoxon signed-rank test comparing the AUC of \texttt{BAG} and \texttt{\resvm} with alternative hypothesis $h_1: AUC^{\resvm} > AUC^{BAG}$ and the number of times each approach had best test set performance.
Test result encoding: $\bullet$ $p < 0.05$, $\bullet\ \bullet$ $p < 0.01$ and $\bullet\ \bullet\ \bullet$ $p < 0.001$.
}
\label{table:semisupervised}
\end{table}


\subsection{Results of semi-supervised classification}
In the semi-supervised setting we deliberately violated the assumption of an uncontaminated positive training set by contaminating $\mathcal{P}$ and $\mathcal{U}$. The results listed in Table~\ref{table:semisupervised} confirm that both class-weighted and bagging SVM are vulnerable to contamination in $\mathcal{P}$ and experience very large performance losses. We believe this is induced by using high misclassification penalties for training instances in $\mathcal{P}$ without any resampling to account for potential false positives. In bagging SVM this leads to a systematic bias in all base models. The resampling strategy of RESVM prevents systematic bias over all base models. 

The results clearly show that RESVM is more robust to false positives, evidenced by a much lower drop in predictive performance for almost all data sets. The performance difference between bagging SVM and RESVM is statistically significant for all data sets except \texttt{covtype} and \texttt{sensit}. Surprisingly, CWSVM obtains 8 wins on \texttt{sensit} with class 2 as positive. RESVM shows the best and most consistent performance overall. 

\begin{figure}[!ht]
  \centering
  \begin{subfigure}[b]{0.48\textwidth}
		\includegraphics[width=\textwidth]{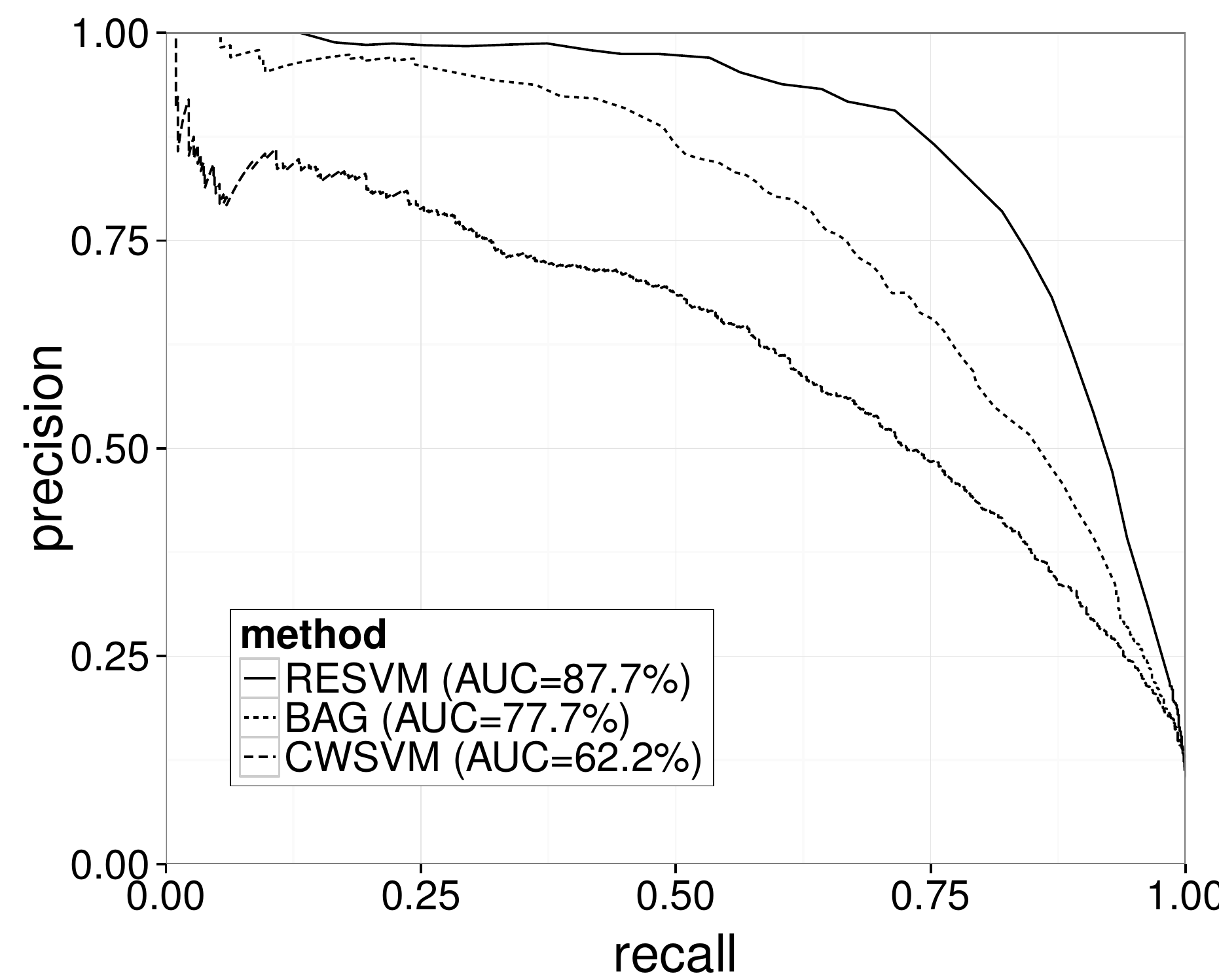}
		\caption{Precision-Recall curves.} \label{fig:pr}
  \end{subfigure}
  ~
  \begin{subfigure}[b]{0.48\textwidth}
	  \includegraphics[width=\textwidth]{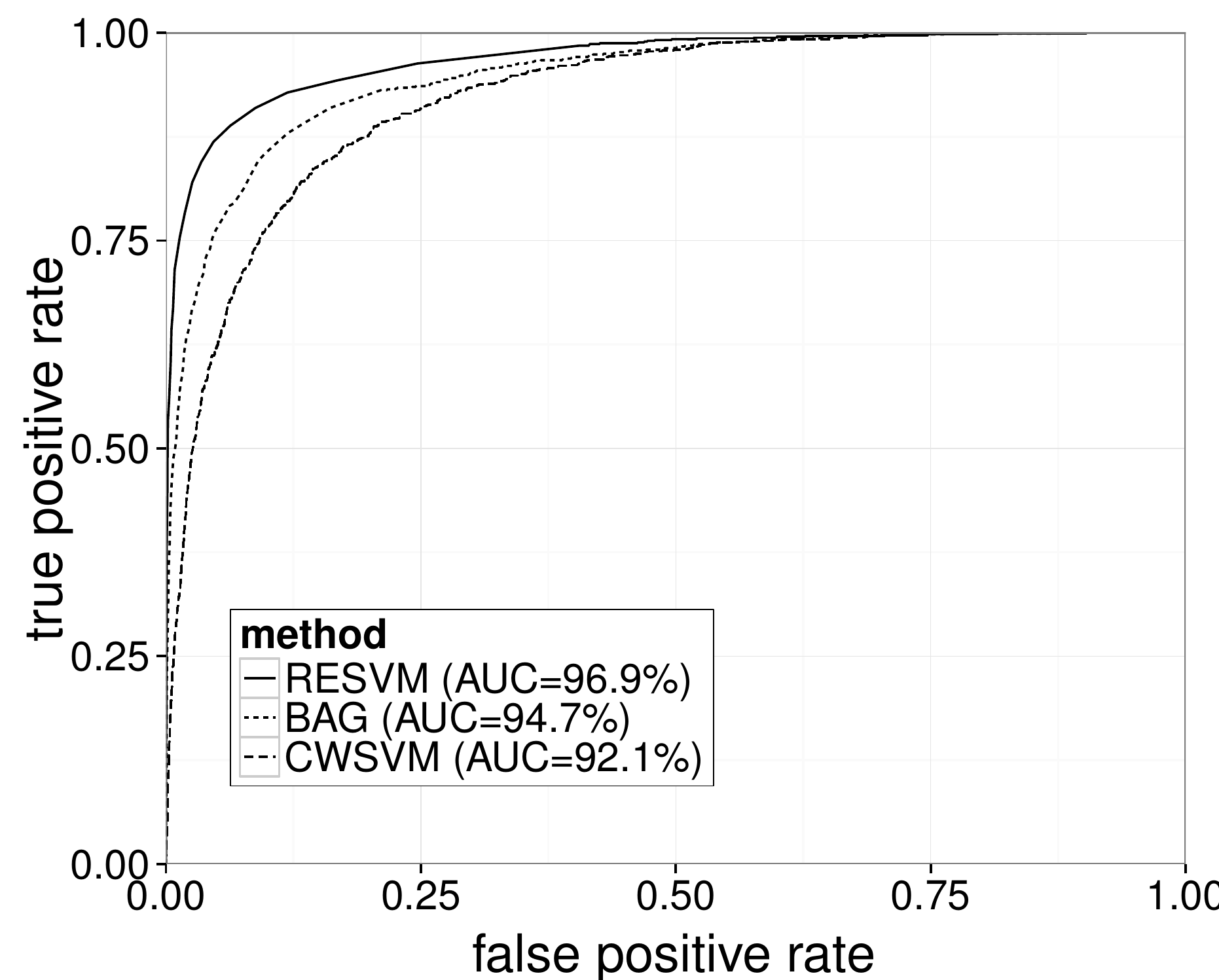}
		\caption{ROC curves.}
  \end{subfigure}
  \caption{Performance in semi-supervised setting on \texttt{mnist}, digit $7$ as positive.} 
  \label{fig:curves}
\end{figure}

On the \texttt{mnist} data, RESVM not only achieved consistently higher area under the PR curve, but visual inspection showed that its PR curves almost always dominated the others over the entire range. This means that in this experiment, RESVM models are always better than the others regardless of design priorities (high precision versus high recall). As an illustration, Figure~\ref{fig:curves} shows the PR and ROC curves of a representative simulation with digit 7 as positive. Since the PR curve of RESVM completely dominates the others we know that its ROC curve does too \citep{Davis:2006:RPR:1143844.1143874}. 

Finally, it is worth noting that the confidence intervals of RESVM tend to be narrower than those of both other approaches. Even though RESVM base models have more variability compared to bagging SVM base models, the overall performance of RESVM is more reliable. This constitutes an important practical advantage since assessing different models is not trivial outside of simulation studies (e.g. when no negative labels are available). 


\subsection{A note on the number of repetitions per experiment}
The tightness of the confidence intervals of generalization performance allow us to conclude that the number of repetitions (20) is sufficient to demonstrate the merits of RESVM (see Tables \ref{table:supervised}--\ref{table:semisupervised}). Increasing the number of repetitions further would yield even narrower confidence intervals and increase the amount of statistically significant results in the Wilcoxon signed-rank test comparing bagging SVM and RESVM (due to increased power). All key conclusions remain valid if the number of repetitions would be increased. 

Additional statistically significant results may only be obtained in experiments where the improvement offered by RESVM is too small to be of practical significance (as large improvements already yield significant test results). Failure to reject the null hypothesis ($h_0: AUC^{BAG} \geq AUC^{RESVM}$) in our current results indicates that (i) bagging SVM is effectively better than RESVM, (ii) they are comparable or (iii) the performance improvement of RESVM is too small to yield a significant test result given the current sample size (number of repetitions). Increasing the number of repetitions can only lead to additional statistically significant results in the latter situation. 

To illustrate our claims, we performed 100 repetitions for \texttt{covtype} in the semi-supervised setting. This yielded the following CIs and win counts: CWSVM $59.0$--$60.5$\% (8 wins), bagging SVM $62.3$--$63.5$\% (21 wins), RESVM $63.8$--$65.8\%$ (71 wins). The $p$-value of the Wilcoxon signed-rank test becomes $2\times 10^{-5}$, while the $p$-value was insignificant with 20 repetitions (Table~\ref{table:semisupervised}).

\subsection{Trend across data sets}
In the previous tables we have shown the results per data set for each setting. In this section we summarize the results across all data sets, using critical difference diagrams \citep{demvsar2006statistical} in Section~\ref{cds} and an overview of win counts in Section~\ref{wins}.

\subsubsection{Critical difference diagrams} \label{cds}
\begin{figure}[!ht]
  \centering
  \begin{subfigure}[b]{0.43\textwidth}
		\includegraphics[width=\textwidth]{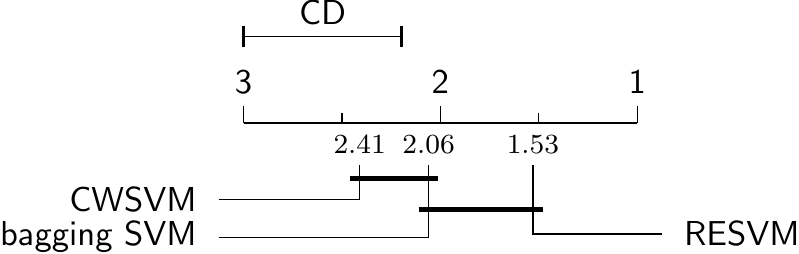}
		\caption{Supervised.} \label{fig:cd_supervised}
  \end{subfigure}
  ~
  \begin{subfigure}[b]{0.43\textwidth}
	  \includegraphics[width=\textwidth]{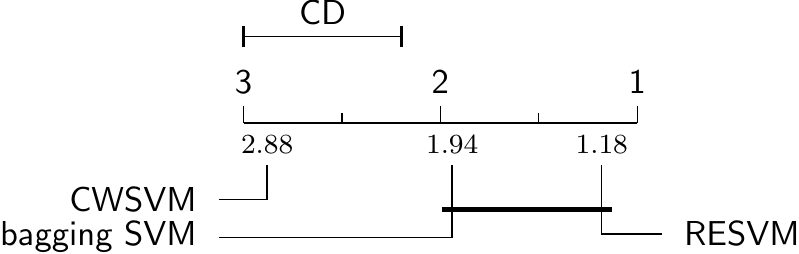}
		\caption{PU learning.} \label{fig:cd_pu}
  \end{subfigure}
  ~
  \begin{subfigure}[b]{0.43\textwidth}
	  \includegraphics[width=\textwidth]{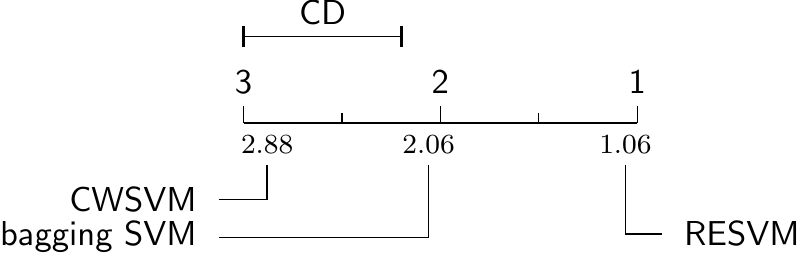}
		\caption{Semi-supervised.} \label{fig:cd_semi}
  \end{subfigure}
  \caption{Critical difference diagrams for each setting. 
      Groups of algorithms that are not significantly different at the $5\%$ significance level are connected.}
  \label{fig:cds}
\end{figure}

In every setting, we compared the performance of the three learning approaches across all data sets using non-parametric statistical tests. For each data set, approaches were ranked based on their mean area under the PR curve across all iterations. Multiclass data sets count once per class. Friedman tests per setting yielded significant evidence of differences between the three learning approaches at the $\alpha=0.05$ level, though this was marginal in the supervised setting ($p=0.034$). The Nemenyi post-hoc test \citep{nemenyi1962distribution} was used after each omnibus test to assess differences between all approaches. The critical difference diagrams in Figure~\ref{fig:cds} visualize the results.

Critical difference diagrams were introduced by \citet{demvsar2006statistical} to visualize a comparison of multiple learning approaches over multiple data sets. These diagrams depict the average rank of each approach (lower is better) along with the critical difference (CD). The critical difference is the minimum difference in average ranks that yields a significant result in the Nemenyi post-hoc test. It depends on the significance level ($\alpha=0.05$), the number of learning approaches ($3$) and the number of data sets ($17$). 

From Figure~\ref{fig:cds} we can conclude that bagging SVM and RESVM are comparable in the PU learning setting (both significantly better than CWSVM). In the semi-supervised setting, bagging SVM is statistically significantly better than CWSVM and RESVM is significantly better than both other approaches across all data sets.

\subsubsection{Win counts} \label{wins}
The number of wins per method across all data sets are summarized in Table~\ref{table:wins}. The top half shows the total number of wins across all data sets, which weights \texttt{mnist} and \texttt{sensit} heavier than the other data sets since we performed several one-vs-all experiments. Because RESVM consistently performed very strong on \texttt{mnist}, the top half is an overly optimistic representation.

The bottom half of Table~\ref{table:wins} contains normalized results, where every data set contributes equally. Based on these numbers we can conclude that there is little difference between the three methods in a supervised setting. In the PU learning setting, ensemble methods become favorable over CWSVM (bagging SVM and RESVM being competitive). Finally, in the semi-supervised setting RESVM pulls far ahead of both other methods and obtains $65\%$ of the normalized wins, which is over three times more than bagging SVM and over five times more than class-weighted SVM.

\begin{table}[!h]
\centering
\begin{tabular}{lcccccccc}
\toprule
    & \multicolumn{2}{c}{CWSVM} & & \multicolumn{2}{c}{bagging SVM} & & \multicolumn{2}{c}{RESVM}\\ \cline{2-3} \cline{5-6} \cline{8-9}
setting         & count & win $\%$ & & count & win $\%$ & & count & win $\%$ \\
\midrule    
supervised      & 73    & 21 & & 93 & 27 & & \textbf{174} & \textbf{51} \\
PU learning     & 21     & 6  & & 88 & 26 & & \textbf{231} & \textbf{68} \\
semi-supervised  & 25    & 7  & & 42 & 12 & & \textbf{273} & \textbf{80}\\
\midrule
supervised      & \textbf{44.8} & \textbf{37.3} & & 40.3 & 33.6 & & 36.0 & 30.0 \\
PU learning     & 18.3  & 15.3 & & 39.4 & 32.8 & & \textbf{62.2} & \textbf{51.8} \\
semi-supervised  & 17.0  & 14.2 & & 24.0 & 20.0 & & \textbf{79.0} & \textbf{65.8} \\
\bottomrule
\end{tabular}
\caption{Number of wins in simulations for each method per setting. The bottom half shows normalized number of wins, where wins in multiclass data sets (\texttt{mnist} and \texttt{sensit}) are divided by the number of classes.}
\label{table:wins}
\end{table}


\subsection{Effect of contamination} \label{varycontamination}
In this Section we show the effect of different levels of contamination in $\mathcal{P}$ and $\mathcal{U}$ on the \texttt{synthetic} data set. In these simulations, we fixed the contamination level in one part of the training set ($\mathcal{P}$ or $\mathcal{U}$) and the contamination of other was varied. The fixed contamination was set to $30\%$. Twenty simulations were run per contamination setting.

In these experiments, we used random search to tune hyperparameters of each method \citep{bergstra2012random} using the Optunity package.\footnote{Optunity is available at: \url{http://www.optunity.net}.} Briefly, hyperparameters were searched by random sampling 100 tuples uniformly within a given box and subsequently the best tuple was selected as before. We ensured that the optimal hyperparameters were never too close to the edge of the feasible region (if so, the box was expanded). Note that this approach of testing a fixed number of tuples favors methods with less hyperparameters. Even though RESVM has more hyperparameters than the other methods, good models can be obtained at the same search cost.

\begin{figure}[!ht]
  \centering
  \begin{subfigure}[b]{0.48\textwidth}
		\includegraphics[width=\textwidth]{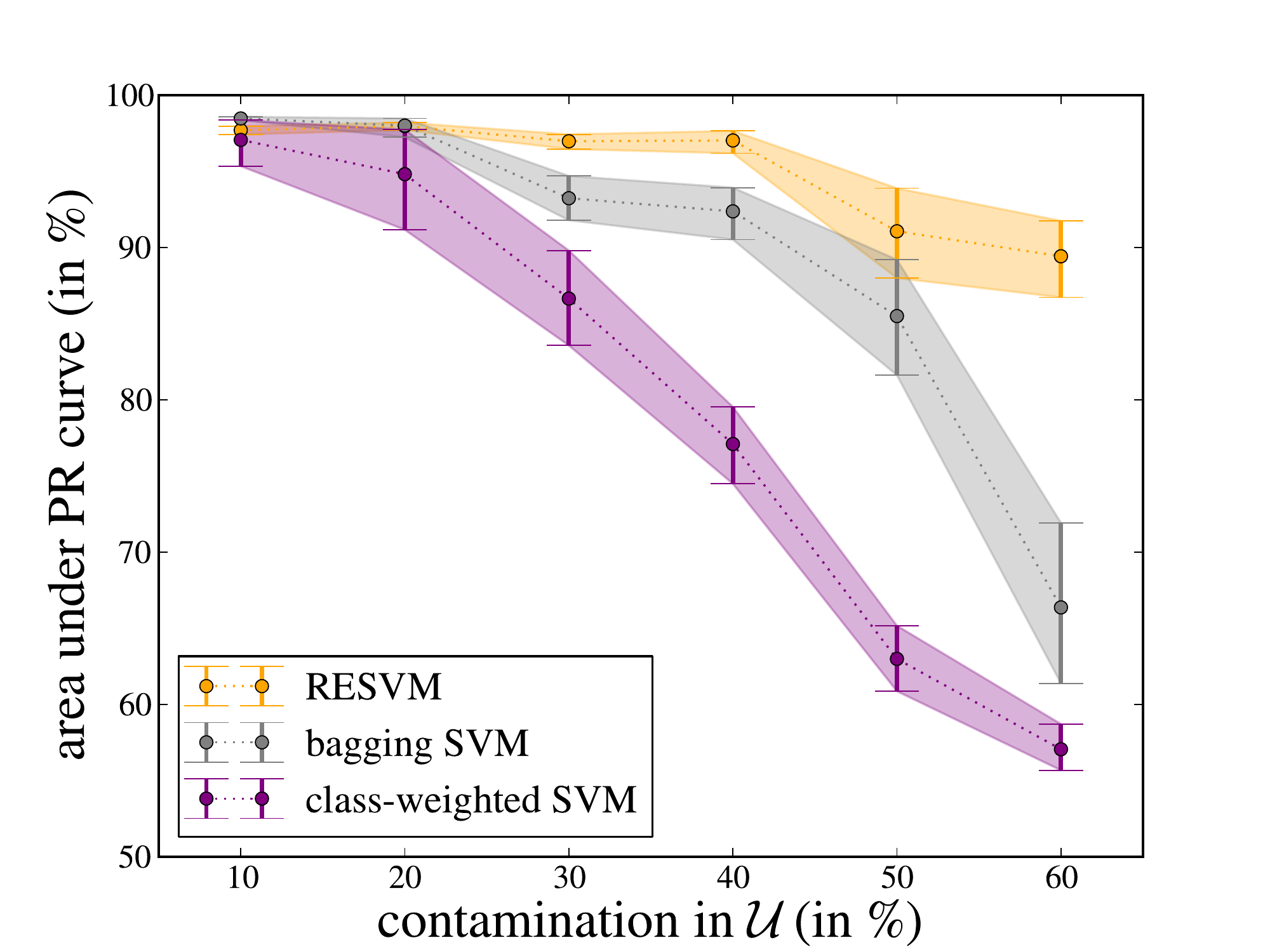}
		\caption{Effect of contamination in $\mathcal{U}$.} \label{fig:contaminationU}
  \end{subfigure}
  ~
  \begin{subfigure}[b]{0.48\textwidth}
	  \includegraphics[width=\textwidth]{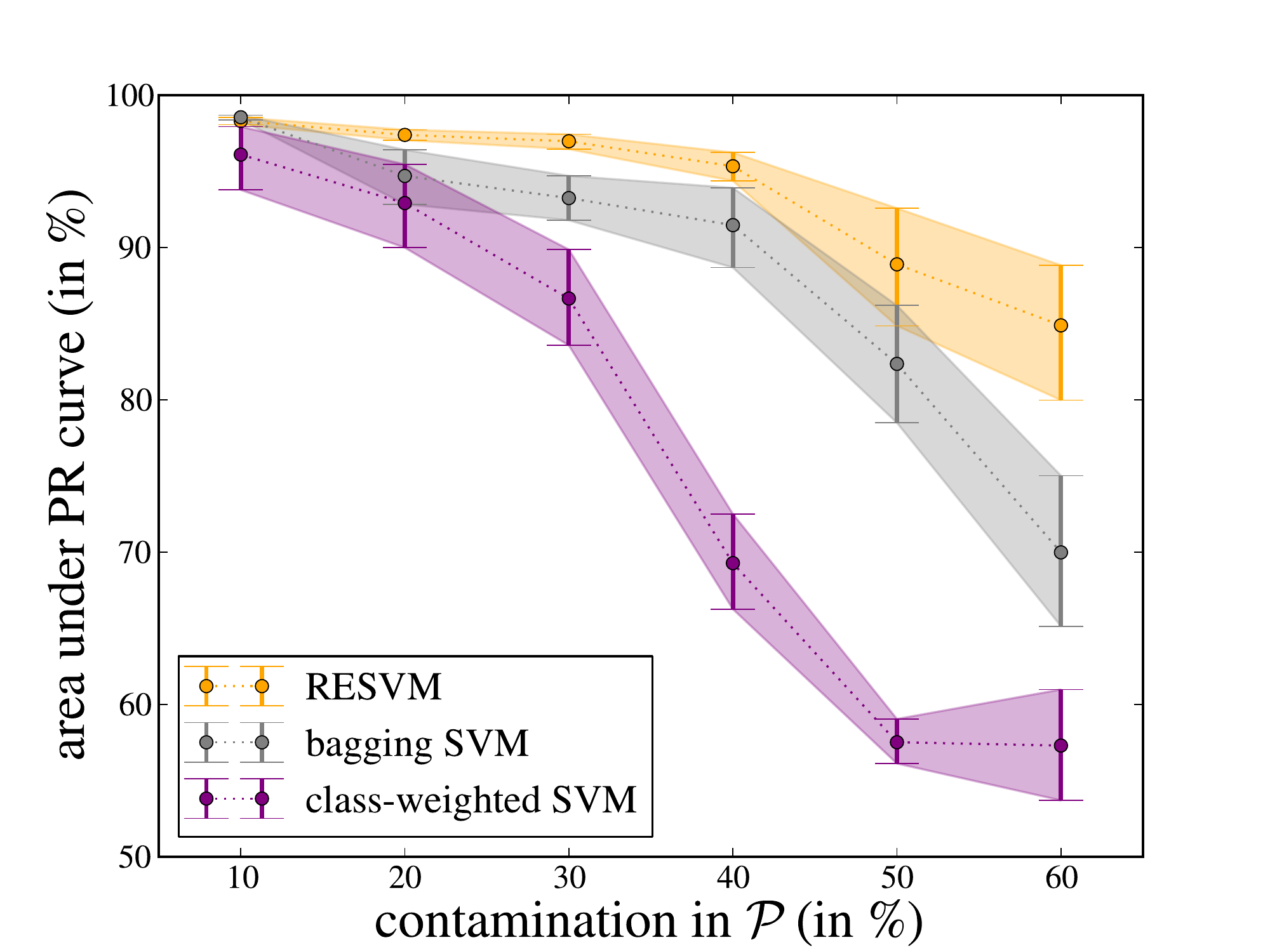}
		\caption{Effect of contamination in $\mathcal{P}$.} \label{fig:contaminationP}
  \end{subfigure}
  \caption{Effect of different levels of contamination in $\mathcal{U}$ and $\mathcal{P}$ on generalization performance. The plots show point estimates of the mean area under the PR curve across experiments and the associated $95\%$ confidence intervals.} 
  \label{fig:contaminationeffect}
\end{figure}

The results are shown in Figure~\ref{fig:contaminationeffect}. In general, contamination in $\mathcal{P}$ causes larger performance losses than the same level of contamination in $\mathcal{U}$ for all algorithms. As expected, the difference in sensitivity to contamination in $\mathcal{P}$ and $\mathcal{U}$ is smallest for RESVM in which $\mathcal{P}$ and $\mathcal{U}$ are resampled similarly. At high contamination levels, RESVM is the only method that still works well (even at $60\%$). 

Figure~\ref{fig:contaminationU} illustrates that RESVM and bagging SVM behave in a similar fashion at contamination levels of $\mathcal{U}$ up to $50\%$ and both outperform class-weighted SVM. RESVM outperforms bagging SVM for contamination levels of $30$--$50\%$ but the consistency (width of CI) and performance losses of both methods are comparable. Figure~\ref{fig:contaminationP} shows the increased robustness of RESVM to contamination in $\mathcal{P}$ resulting in reduced loss of generalization performance for increasing contamination. 

\subsection{RESVM optimal parameters}
As an illustration of the implicit mechanism of RESVM we show some of the optimal tuning parameters for every setting in Table~\ref{table:hyperparameters}. These parameters were obtained by performing 10-fold cross-validation on the training set. 

\begin{table}[!h]
\centering
\begin{tabular}{lccccccccccc}
\toprule
 & $0$ & $1$ & $2$ & $3$ & $4$ & $5$ & $6$ & $7$ & $8$ & $9$ & mean \\
\midrule
$\mathbf{n_{pos}}$ \\
\ \ supervised &   20 &   20 &   20 &   20 &   20 &   20 &   10 &   20 &   20 &   10 &   18 \\
\ \ PU learning &   10 &   10 &   10 &   10 &   10 &   15 &   10 &   10 &   10 &   10 &   10.5 \\
\ \ semi-superv. &   10 &    5 &   10 &   10 &   10 &   10 &   10 &   10 &   10 &   10 &   9.5 \\
$\mathbf{n_{unl}/n_{pos}}$ \\
\ \ supervised  &   10 &   10 &   10 &   10 &   10 &   10 &   10 &   10 &   10 &   10 &   10 \\
\ \ PU learning &    5 &    5 &    5 &    5 &    5 &    5 &    5 &    5 &    5 &    5 &    5 \\
\ \ semi-superv.  &    5 &    5 &    5 &    8 &    5 &    5 &    5 &    5 &    5 &    5 & 5.25  \\
$\mathbf{w_{pos}}$ \\
\ \ supervised  &  1.6 &  1.6 &  1.6 &  3.2 &  3.2 &  3.2 &  3.2 &  1.6 &  3.2 &  2.4 &  2.48 \\
\ \ PU learning &  4.8 &  6.4 &  3.2 &  6.4 &  4.8 &  6.4 &  4.8 &  4.8 &  6.4 &  6.4 &  5.44 \\
\ \ semi-superv.  & 12.8 &  6.4 &  4.8 &  2.1 &  4.8 &  6.4 &  4.8 &  3.2 &  3.2 &  3.2 &  5.17 \\
\bottomrule
\end{tabular}
\caption{Medians of optimal hyperparameters per digit obtained via cross-validation and mean of all medians per setting. The normalized relative weight on positives versus unlabeled instances ($w_{pos}$) is associated with the relative size and contamination of the positive and unlabeled training sets.}
\label{table:hyperparameters}
\end{table}

An interesting observation is that the size of the training sets that are being used decreases for increasing contamination. Increasing label noise induces RESVM to favor smaller base model training sets for which the variability in contamination is larger (see Figure~\ref{fig:contamination}). Though this may appear counterintuitive, bagging approaches are known to exhibit a bias-variance tradeoff \citep{bauer1999empirical} for which using weaker base models with increased variability may yield better ensembles \citep{biasvariance}.

The optimal value of the misclassification penalty for positive training instances relative to unlabeled instances, $w_{pos}$, changes between learning settings (see Equation~\eqref{eq:resvmpenalties}). It exhibits expected behaviour: the maximum value is obtained when the certainty on $\mathcal{P}$ relative to $\mathcal{U}$ is largest (e.g. the pure PU learning setting). This parameter implicitly balances empirical certainty on $\mathcal{P}$ and $\mathcal{U}$ and is an important degree of freedom in RESVM. In bagging SVM, this parameter is implicitly fixed to $1$ via Equation~\eqref{eq:bagpenalties} \citep{MORDELET-2010-523336}. Note that $w_{pos}$ need not be larger than 1 (which would place extra emphasisis on the known labels after accounting for class imbalance). In highly imbalanced settings where $n_{unl} \gg n_{pos}$, the optimal value of $w_{pos}$ may well be less than 1.


\section{Conclusion}
We have introduced a new approach for learning from positive and unlabeled data, called the robust ensemble of SVMs (RESVM). RESVM constructs an ensemble model using a bagging strategy in which the positive and unlabeled sets are resampled to obtain base model training sets. By resampling both $\mathcal{P}$ and $\mathcal{U}$, our approach is more robust against false positives than others.

The robustness of our approach to potential contamination in both $\mathcal{P}$ and $\mathcal{U}$ can be attributed to the synergy between our resampling scheme and voting aggregation. The resampling itself strongly resembles a typical bootstrap approach. RESVM uses class-weighted SVM base models though the resampling scheme is likely to work well with other types of base models.

RESVM was compared with class-weighted SVM and bagging SVM on several data sets under different label noise conditions. The trends across data sets show that bagging SVM and RESVM outperform class-weighted SVM in PU learning. In a pure PU learning setting the average improvement over existing methods is modest though RESVM classifiers exhibit lower variance in performance making it more reliable.

In the semi-supervised setting, label noise was introduced in $\mathcal{P}$ to highlight the improved robustness of RESVM compared to the other methods. Our experimental results show that RESVM remains very strong in the semi-supervised setting while both other approaches degrade dramatically. Statistical analysis showed that RESVM is significantly better than both other approaches across all data sets.



Visual inspection of the PR curves shows that in the majority of experiments the curve for RESVM not only has higher AUC but completely dominates the other curves. As such RESVM models are a good approach regardless of design priorities (high recall versus high precision).

A weakness of RESVM is its amount of hyperparameters ($5$ plus potential kernel parameters), though RESVM models are less sensitive to accurate tuning of these parameters than standard SVM. Our experiments indicated that although RESVM has more hyperparameters, good models can be obtained at the same search effort than the other approaches (e.g. testing the same number of hyperparameter tuples). An interesting question is whether prior knowledge regarding contamination of $\mathcal{P}$ and $\mathcal{U}$ can help in limiting the search scope for some of the hyperparameters ($n_{pos}$, $n_{unl}$ and $w_{pos}$ specifically).


\section*{Acknowledgements}
We wish to thank the anonymous reviewers for their valuable comments which helped us to improve the quality of the manuscript.

This research was supported by: 
\begin{itemize}
\item Research Council KU Leuven: GOA/10/09 MaNet, CoE PFV/10/002 (OPTEC), BIL12/11T; PhD/Postdoc grants
\item Flemish Government:
\begin{itemize}
\item FWO: projects:  G.0871.12N (Neural circuits), G.0377.12 (Structured systems), G.088114N (Tensor based data similarity); PhD/Postdoc grant
\item IWT: TBM-Logic Insulin(100793), TBM Rectal Cancer(100783), TBM IETA(130256), POM II  SBO 100031; PhD grant number 111065
\item Industrial Research fund (IOF): IOF/HB/13/027 Logic Insulin
\item iMinds Medical IT SBO 2014
\item VLK Stichting E. van der Schueren: rectal cancer
\end{itemize}
\item Federal Government: FOD: Cancer Plan 2012-2015 KPC-29-023 (prostate), Belgian Federal Science Policy Office: IUAP P7/19 (DYSCO, Dynamical systems, control and optimization, 2012-2017)
\item COST: Action: BM1104: Mass Spectrometry Imaging
\item EU: The research leading to these results has received funding from the European Research Council under the European Union's Seventh Framework Programme (FP7/2007-2013) / ERC AdG A-DATADRIVE-B (290923).  This paper reflects only the authors' views and the Union is not liable for any use that may be made of the contained information.
\end{itemize}

\section*{References}
\bibliographystyle{plainnat}
\bibliography{bibliography}

\begin{thebibliography}{45}
\providecommand{\natexlab}[1]{#1}
\providecommand{\url}[1]{\texttt{#1}}
\expandafter\ifx\csname urlstyle\endcsname\relax
  \providecommand{\doi}[1]{doi: #1}\else
  \providecommand{\doi}{doi: \begingroup \urlstyle{rm}\Url}\fi

\bibitem[Aerts et~al.(2006)Aerts, Lambrechts, Maity, Van~Loo, Coessens,
  De~Smet, Tranchevent, De~Moor, Marynen, Hassan, Carmeliet, and
  Moreau]{citeulike:615632}
Stein Aerts, Diether Lambrechts, Sunit Maity, Peter Van~Loo, Bert Coessens,
  Frederik De~Smet, Leon-Charles Tranchevent, Bart De~Moor, Peter Marynen,
  Bassem Hassan, Peter Carmeliet, and Yves Moreau.
\newblock {Gene prioritization through genomic data fusion}.
\newblock \emph{Nature Biotechnology}, 24\penalty0 (5):\penalty0 537--544, May
  2006.
\newblock ISSN 1087-0156.

\bibitem[Alzate and Suykens(2012)]{bksc2}
Carlos Alzate and Johan A.~K. Suykens.
\newblock A semi-supervised formulation to binary kernel spectral clustering.
\newblock In \emph{2012 IEEE World Congress on Computational Intelligence (IEEE
  WCCI/IJCNN 2012)}, Brisbane, Australia, June 2012.

\bibitem[Banfield et~al.(2007)Banfield, Hall, Bowyer, and
  Kegelmeyer]{banfield2007comparison}
Robert~E Banfield, Lawrence~O Hall, Kevin~W Bowyer, and W~Philip Kegelmeyer.
\newblock A comparison of decision tree ensemble creation techniques.
\newblock \emph{Pattern Analysis and Machine Intelligence, IEEE Transactions
  on}, 29\penalty0 (1):\penalty0 173--180, 2007.

\bibitem[Bauer and Kohavi(1999)]{bauer1999empirical}
Eric Bauer and Ron Kohavi.
\newblock An empirical comparison of voting classification algorithms: Bagging,
  boosting, and variants.
\newblock \emph{Machine learning}, 36\penalty0 (1-2):\penalty0 105--139, 1999.

\bibitem[Bergstra and Bengio(2012)]{bergstra2012random}
James Bergstra and Yoshua Bengio.
\newblock Random search for hyper-parameter optimization.
\newblock \emph{Journal of Machine Learning Research}, 13\penalty0
  (1):\penalty0 281--305, 2012.

\bibitem[Blackard and Dean(1999)]{Blackard00covtype}
Jock~A. Blackard and Denis~J. Dean.
\newblock {Comparative accuracies of artificial neural networks and
  discriminant analysis in predicting forest cover types from cartographic
  variables}.
\newblock \emph{Computers and Electronics in Agriculture}, 24\penalty0
  (3):\penalty0 131--151, December 1999.

\bibitem[Blanchard et~al.(2010)Blanchard, Lee, and Scott]{blanchard2010semi}
Gilles Blanchard, Gyemin Lee, and Clayton Scott.
\newblock Semi-supervised novelty detection.
\newblock \emph{Journal of Machine Learning Research}, 11:\penalty0 2973--3009,
  2010.

\bibitem[Bottou and Lin(2007)]{bottou2007support}
L{\'e}on Bottou and Chih-Jen Lin.
\newblock Support vector machine solvers.
\newblock \emph{Large scale kernel machines}, pages 301--320, 2007.

\bibitem[Breiman(1996)]{Breiman:1996:BP:231986.231989}
Leo Breiman.
\newblock Bagging predictors.
\newblock \emph{Machine Learning}, 24\penalty0 (2):\penalty0 123--140, August
  1996.
\newblock ISSN 0885-6125.

\bibitem[Breiman(2000)]{breiman2000randomizing}
Leo Breiman.
\newblock Randomizing outputs to increase prediction accuracy.
\newblock \emph{Machine Learning}, 40\penalty0 (3):\penalty0 229--242, 2000.

\bibitem[Breiman(2001)]{breiman2001random}
Leo Breiman.
\newblock Random forests.
\newblock \emph{Machine learning}, 45\penalty0 (1):\penalty0 5--32, 2001.

\bibitem[Brown et~al.(2005)Brown, Wyatt, Harris, and Yao]{brown2005diversity}
Gavin Brown, Jeremy Wyatt, Rachel Harris, and Xin Yao.
\newblock Diversity creation methods: a survey and categorisation.
\newblock \emph{Information Fusion}, 6\penalty0 (1):\penalty0 5--20, 2005.

\bibitem[Cawley(2006)]{cawley2006leave}
Gavin~C Cawley.
\newblock Leave-one-out cross-validation based model selection criteria for
  weighted {LS-SVMs}.
\newblock In \emph{Neural Networks, 2006. IJCNN'06. International Joint
  Conference on}, pages 1661--1668. IEEE, 2006.

\bibitem[Chang and Lin(2011)]{CC01a}
Chih-Chung Chang and Chih-Jen Lin.
\newblock {LIBSVM}: A library for support vector machines.
\newblock \emph{ACM Transactions on Intelligent Systems and Technology},
  2:\penalty0 27:1--27:27, 2011.
\newblock Software available at \url{http://www.csie.ntu.edu.tw/~cjlin/libsvm}.

\bibitem[Claesen et~al.(2014)Claesen, Smet, Suykens, and Moor]{claesen14}
Marc Claesen, Frank~De Smet, Johan~A.K. Suykens, and Bart~De Moor.
\newblock {EnsembleSVM}: A library for ensemble learning using support vector
  machines.
\newblock \emph{Journal of Machine Learning Research}, 15:\penalty0 141--145,
  2014.
\newblock URL \url{http://jmlr.org/papers/v15/claesen14a.html}.

\bibitem[Daemen et~al.(2009)Daemen, Gevaert, Ojeda, Debucquoy, Suykens,
  Sempoux, Machiels, Haustermans, and De~Moor]{daemen2009kernel}
Anneleen Daemen, Olivier Gevaert, Fabian Ojeda, Annelies Debucquoy, Johan~A.K.
  Suykens, Christine Sempoux, Jean-Pascal Machiels, Karin Haustermans, and Bart
  De~Moor.
\newblock A kernel-based integration of genome-wide data for clinical decision
  support.
\newblock \emph{Genome Medicine}, 1\penalty0 (4):\penalty0 39, 2009.

\bibitem[Davis and Goadrich(2006)]{Davis:2006:RPR:1143844.1143874}
Jesse Davis and Mark Goadrich.
\newblock The relationship between precision-recall and {ROC} curves.
\newblock In \emph{Proceedings of the 23rd international conference on Machine
  learning}, ICML '06, pages 233--240, New York, NY, USA, 2006. ACM.
\newblock ISBN 1-59593-383-2.

\bibitem[Dem{\v{s}}ar(2006)]{demvsar2006statistical}
Janez Dem{\v{s}}ar.
\newblock Statistical comparisons of classifiers over multiple data sets.
\newblock \emph{Journal of Machine Learning Research}, 7:\penalty0 1--30, 2006.

\bibitem[Dietterich(2000)]{dietterich2000experimental}
Thomas~G Dietterich.
\newblock An experimental comparison of three methods for constructing
  ensembles of decision trees: Bagging, boosting, and randomization.
\newblock \emph{Machine learning}, 40\penalty0 (2):\penalty0 139--157, 2000.

\bibitem[Duarte and Hen~Hu(2004)]{duarte2004vehicle}
Marco~F Duarte and Yu~Hen~Hu.
\newblock Vehicle classification in distributed sensor networks.
\newblock \emph{Journal of Parallel and Distributed Computing}, 64\penalty0
  (7):\penalty0 826--838, 2004.

\bibitem[Elkan and Noto(2008)]{Elkan:2008:LCO:1401890.1401920}
Charles Elkan and Keith Noto.
\newblock Learning classifiers from only positive and unlabeled data.
\newblock In \emph{Proceedings of the 14th ACM SIGKDD international conference
  on Knowledge discovery and data mining}, KDD '08, pages 213--220, New York,
  NY, USA, 2008. ACM.

\bibitem[Fan et~al.(2008)Fan, Chang, Hsieh, Wang, and Lin]{fan2008liblinear}
Rong-En Fan, Kai-Wei Chang, Cho-Jui Hsieh, Xiang-Rui Wang, and Chih-Jen Lin.
\newblock {LIBLINEAR}: A library for large linear classification.
\newblock \emph{Journal of Machine Learning Research}, 9:\penalty0 1871--1874,
  2008.

\bibitem[Frenay and Verleysen(2014)]{frenay}
B.~Frenay and M.~Verleysen.
\newblock Classification in the presence of label noise: A survey.
\newblock \emph{Neural Networks and Learning Systems, IEEE Transactions on},
  25\penalty0 (5):\penalty0 845--869, May 2014.
\newblock ISSN 2162-237X.
\newblock \doi{10.1109/TNNLS.2013.2292894}.

\bibitem[Grandvalet(2004)]{grandvalet2004bagging}
Yves Grandvalet.
\newblock Bagging equalizes influence.
\newblock \emph{Machine Learning}, 55\penalty0 (3):\penalty0 251--270, 2004.

\bibitem[Keijzer and Babovic(2000)]{biasvariance}
Maarten Keijzer and Vladan Babovic.
\newblock Genetic programming, ensemble methods and the bias/variance tradeoff
  – introductory investigations.
\newblock In Riccardo Poli, Wolfgang Banzhaf, William~B. Langdon, Julian
  Miller, Peter Nordin, and Terence~C. Fogarty, editors, \emph{Genetic
  Programming}, volume 1802 of \emph{Lecture Notes in Computer Science}, pages
  76--90. Springer Berlin Heidelberg, 2000.
\newblock ISBN 978-3-540-67339-2.
\newblock \doi{10.1007/978-3-540-46239-2_6}.
\newblock URL \url{http://dx.doi.org/10.1007/978-3-540-46239-2_6}.

\bibitem[Lazarevic et~al.(2003)Lazarevic, Ozgur, Ertoz, Srivastava, and
  Kumar]{Lazarevic03acomparative}
Ar~Lazarevic, Aysel Ozgur, Levent Ertoz, Jaideep Srivastava, and Vipin Kumar.
\newblock A comparative study of anomaly detection schemes in network intrusion
  detection.
\newblock In \emph{Proceedings of the Third SIAM International Conference on
  Data Mining}, 2003.

\bibitem[{LeCun} et~al.(1998){LeCun}, Bottou, Bengio, and
  Haffner]{Lecun98gradient-basedlearning}
Yann {LeCun}, L\'eon Bottou, Yoshua Bengio, and Patrick Haffner.
\newblock Gradient-based learning applied to document recognition.
\newblock In \emph{Proceedings of the IEEE}, pages 2278--2324, 1998.

\bibitem[Lee and Liu(2003)]{Lee03learningwith}
Wee~Sun Lee and Bing Liu.
\newblock Learning with positive and unlabeled examples using weighted logistic
  regression.
\newblock In \emph{Proceedings of the Twentieth International Conference on
  Machine Learning (ICML)}, pages 448--455, 2003.

\bibitem[Li and Liu(2003)]{Li03learningto}
Xiaoli Li and Bing Liu.
\newblock {Learning to classify texts using positive and unlabeled data}.
\newblock In \emph{IJCAI'03: Proceedings of the 18th international joint
  conference on Artificial intelligence}, pages 587--592, San Francisco, CA,
  USA, 2003. Morgan Kaufmann Publishers Inc.

\bibitem[Liu et~al.(2002)Liu, Lee, Yu, and Li]{liu02partially}
Bing Liu, Wee~Sun Lee, Philip~S. Yu, and Xiaoli Li.
\newblock Partially supervised classification of text documents.
\newblock In \emph{ICML '02: Proceedings of the Nineteenth International
  Conference on Machine Learning}, pages 387--394, San Francisco, CA, USA,
  2002. Morgan Kaufmann Publishers Inc.
\newblock ISBN 1-55860-873-7.

\bibitem[Liu et~al.(2003)Liu, Dai, Li, Lee, and Yu]{Liu:2003:BTC:951949.952139}
Bing Liu, Yang Dai, Xiaoli Li, Wee~Sun Lee, and Philip~S. Yu.
\newblock Building text classifiers using positive and unlabeled examples.
\newblock In \emph{Proceedings of the Third IEEE International Conference on
  Data Mining}, ICDM '03, pages 179--186, Washington, DC, USA, 2003. IEEE
  Computer Society.
\newblock ISBN 0-7695-1978-4.

\bibitem[Liu et~al.(2005)Liu, Shi, Li, and Qin]{Liu:2005:PSC:2138033.2138052}
Zhigang Liu, Wenzhong Shi, Deren Li, and Qianqing Qin.
\newblock Partially supervised classification -- based on weighted unlabeled
  samples support vector machine.
\newblock In \emph{Proceedings of the First international conference on
  Advanced Data Mining and Applications}, ADMA'05, pages 118--129, Berlin,
  Heidelberg, 2005. Springer-Verlag.
\newblock ISBN 3-540-27894-X, 978-3-540-27894-8.

\bibitem[Mart{\'\i}nez-Mu{\~n}oz and Su{\'a}rez(2010)]{martinez2010out}
Gonzalo Mart{\'\i}nez-Mu{\~n}oz and Alberto Su{\'a}rez.
\newblock Out-of-bag estimation of the optimal sample size in bagging.
\newblock \emph{Pattern Recognition}, 43\penalty0 (1):\penalty0 143--152, 2010.

\bibitem[Mordelet and Vert(2010)]{MORDELET-2010-523336}
Fantine Mordelet and Jean-Philippe Vert.
\newblock A bagging {SVM} to learn from positive and unlabeled examples.
\newblock \emph{arXiv preprint arXiv:1010.0772}, 2010.

\bibitem[Mordelet and Vert(2011)]{mordelet2011prodige}
Fantine Mordelet and Jean-Philippe Vert.
\newblock {ProDiGe}: Prioritization of disease genes with multitask machine
  learning from positive and unlabeled examples.
\newblock \emph{BMC bioinformatics}, 12\penalty0 (1):\penalty0 389, 2011.

\bibitem[Nemenyi(1962)]{nemenyi1962distribution}
Peter Nemenyi.
\newblock Distribution-free multiple comparisons.
\newblock In \emph{BIOMETRICS}, volume~18, page 263. INTERNATIONAL BIOMETRIC
  SOC 1441 I ST, NW, SUITE 700, WASHINGTON, DC 20005-2210, 1962.

\bibitem[Osuna et~al.(1997)Osuna, Freund, and Girosi]{osuna1997}
Edgar Osuna, Robert Freund, and Federico Girosi.
\newblock {Support Vector Machines: Training and Applications}.
\newblock Technical Report AIM-1602, 1997.

\bibitem[Pechenizkiy et~al.(2006)Pechenizkiy, Tsymbal, Puuronen, and
  Pechenizkiy]{pechenizkiy2006class}
Mykola Pechenizkiy, Alexey Tsymbal, Seppo Puuronen, and Oleksandr Pechenizkiy.
\newblock Class noise and supervised learning in medical domains: The effect of
  feature extraction.
\newblock In \emph{Computer-Based Medical Systems, 2006. CBMS 2006. 19th IEEE
  International Symposium on}, pages 708--713. IEEE, 2006.

\bibitem[Prokhorov(2001)]{prokhorov2001ijcnn}
Danil Prokhorov.
\newblock {IJCNN} 2001 neural network competition.
\newblock \emph{Slide presentation in IJCNN}, 2001.

\bibitem[Rao and Tibshirani(1997)]{rao1997out}
J~Sunil Rao and Robert Tibshirani.
\newblock The out-of-bootstrap method for model averaging and selection.
\newblock \emph{University of Toronto}, 1997.

\bibitem[Shoichet(2004)]{citeulike:3911}
Brian~K. Shoichet.
\newblock {Virtual screening of chemical libraries}.
\newblock \emph{Nature}, 432\penalty0 (7019):\penalty0 862--865, December 2004.
\newblock ISSN 0028-0836.

\bibitem[Sifrim et~al.(2013)Sifrim, Popovic, Tranchevent, Arderschirdavani,
  Sakai, Konings, Vermeesch, Aerts, De~Moor, and Moreau]{sifrim2013extasy}
Alejandro Sifrim, Dusan Popovic, L{\'e}on-Charles Tranchevent, Amin
  Arderschirdavani, Ryo Sakai, Peter Konings, Joris Vermeesch, Jan Aerts, Bart
  De~Moor, and Yves Moreau.
\newblock {eXtasy}: Variant prioritization by genomic data fusion.
\newblock \emph{Nature Methods}, 10:\penalty0 1083--1084, 2013.
\newblock \doi{http://dx.doi.org/10.1038/nmeth.2656}.

\bibitem[Yu(2005)]{Yu:2005:SCM:1108759.1108762}
Hwanjo Yu.
\newblock Single-class classification with mapping convergence.
\newblock \emph{Machine Learning}, 61\penalty0 (1-3):\penalty0 49--69, November
  2005.
\newblock ISSN 0885-6125.

\bibitem[Yu et~al.(2002)Yu, Han, and Chang]{Yu02pebl:positive}
Hwanjo Yu, Jiawei Han, and Kevin~C. Chang.
\newblock {PEBL}: positive example based learning for web page classification
  using {SVM}.
\newblock In \emph{KDD '02: Proceedings of the eighth ACM SIGKDD international
  conference on Knowledge discovery and data mining}, pages 239--248, New York,
  NY, USA, 2002. ACM Press.
\newblock ISBN 158113567X.

\bibitem[Zhu and Wu(2004)]{zhu2004class}
Xingquan Zhu and Xindong Wu.
\newblock Class noise vs. attribute noise: A quantitative study.
\newblock \emph{Artificial Intelligence Review}, 22\penalty0 (3):\penalty0
  177--210, 2004.

\end{thebibliography}


\section{Vitae}
{\noindent}\textbf{Marc Claesen} was born in Diepenbeek, Belgium on April 5, 1987. He received his Masters degree in Mathematical Engineering in 2010 at KU Leuven, Belgium. He is currently a doctoral student at the same university at the department of Electrical Engineering (ESAT) in the STADIUS research group. His research interests include machine learning, open-source software, kernel methods, large-scale and semi-supervised learning. Further information, including an updated CV, can be found at \url{www.marc-claesen.name}. \\

{\noindent}\textbf{Frank De Smet} was born in Bonheiden, Belgium, in August 1969. He received the M.S. degree in electrical and mechanical engineering and is a Medical Doctor from the KU Leuven (Belgium) in 1992 and 1998, respectively. He obtained his PhD in electrical engineering (bioinformatics) from the same university in 2004. Currently he is a visiting professor at the department of Public Health and Primary Care of the KU Leuven. He is also a member of the medical management department of the National Alliance of Christian Mutualities where he focusses, among others, on insurance medicine and public health, data mining and fraud detection, eHealth, quality in healthcare, patient empowerment, and evidence-based medicine (EBM).\\

{\noindent}\textbf{Johan A.K. Suykens} was born in Willebroek Belgium, May 18, 1966. He received the degree in Electro-Mechanical Engineering and the Ph.D. degree in Applied Sciences from the Katholieke Universiteit Leuven, in 1989 and 1995, respectively. In 1996 he has been a Visiting Postdoctoral Researcher at the University of California, Berkeley. He has been a Postdoctoral Researcher with the Fund for Scientific Research FWO Flanders and is currently a Professor (Hoogleraar) with KU Leuven. He is the author of the books ``Artificial Neural Networks for Modelling and Control of Non-linear Systems" (Kluwer Academic Publishers) and ``Least Squares Support Vector Machines" (World Scientific), co-author of the book ``Cellular Neural Networks, Multi-Scroll Chaos and Synchronization" (World Scientific) and editor of the books ``Nonlinear Modeling: Advanced Black-Box Techniques" (Kluwer Academic Publishers) and ``Advances in Learning Theory: Methods, Models and Applications" (IOS Press). In 1998 he organized an International Workshop on Nonlinear Modelling with Time-series Prediction Competition. He is a Senior IEEE member and has served as an associate editor for the IEEE Transactions on Circuits and Systems (1997--1999 and 2004--2007) and for the IEEE Transactions on Neural Networks (1998--2009). He received an IEEE Signal Processing Society 1999 Best Paper (Senior) Award and several Best Paper Awards at International Conferences. He is a recipient of the International Neural Networks Society INNS 2000 Young Investigator Award for significant contributions in the field of neural networks. He has served as a Director and Organizer of the NATO Advanced Study Institute on Learning Theory and Practice (Leuven 2002), as a program co-chair for the International Joint Conference on Neural Networks 2004 and the International Symposium on Nonlinear Theory and its Applications 2005, as an organizer of the International Symposium on Synchronization in Complex Networks 2007 and a co-organizer of the NIPS 2010 workshop on Tensors, Kernels and Machine Learning. He has been recently awarded an ERC Advanced Grant in 2011. \\

{\noindent}\textbf{Bart De Moor} was born on Tuesday, July 12, 1960, in Halle, Belgium. He is married and has three children. In 1983, he obtained his Master Degree in Electrical Engineering at the KU Leuven, Belgium, and a Ph.D. in Engineering at the same university in 1988. He spent 2 years as a Visiting Research Associate at Stanford University (1988--1990) at the departments of EE (ISL, Professor Kailath) and CS (Professor Golub). Currently, he is a full professor at the Department of Electrical Engineering in the research group STADIUS and the Scientific Director of the iMinds Future Health Department. His research interests are in numerical linear algebra, algebraic geometry and optimization, system theory and system identification, quantum information theory, control theory, data-mining, information retrieval and bio-informatics (for books and research publications, see the publication search engine at \url{http://homes.esat.kuleuven.be/~sistawww/cgi-bin/pub.pl}). He is or has been the coordinator of numerous research projects and networks funded by regional, federal and European funding agencies. Currently, he is leading a research group of 10 Ph.D. students and 4 postdocs and in the recent past, about 80 Ph.D.s were obtained under his guidance. He has been teaching at several universities in Europe and the US. He is a member of several scientific and professional organizations, jury member of several scientific and industrial awards, and chairman or member of international educational and scientific review and selection committees. He is an associate editor of and reviewer for several scientific journals. His work has won him several scientific awards (Leybold-Heraeus Prize (1986), Leslie Fox Prize (1989), Guillemin-Cauer best paper Award of the IEEE Transactions on Circuits and Systems (1990), Laureate of the Belgian Royal Academy of Sciences (1992), biannual Siemens Award (1994), best paper award of Automatica (IFAC, 1996), IEEE Signal Processing Society Best Paper Award (1999)). In November 2010, he received the 5-annual FWO Excellence Award out of the hands of King Albert II of Belgium. Since 2004, he is a fellow of the IEEE (www.ieee.org). From 1991 to 1999 he was the Head of Cabinet and/or Main Advisor on Science and Technology of several ministers of the Belgian Federal Government (Demeester, Martens) and the Flanders Regional Governments (Demeester, Van den Brande). From December 2005 to July 2007, he was the Head of Cabinet on socio-economic policy of the minister-president of Flanders, Yves Leterme, capacity in which he was the coordinator of a new socio-economic business plan for the Flemish region (\url{www.vlaandereninactie.be}). He co-founded 6 spinoff companies, 4 of which are still active (\url{www.ipcos.be}, \url{www.tmleuven.be}, \url{www.dsquare.be}, \url{www.cartagenia.com}). He is in the board of the Flemish Interuniversity Institute for Biotechnology (\url{www.vib.be}), the Study Center for Nuclear Energy (\url{www.sck.be}), the Institute for Broad Band Technology (\url{www.iminds.be}), the Flemish Children Science Center Technopolis (\url{www.technopolis.be}), the Alamire Foundation (\url{http://www.alamirefoundation.org/}) and several other scientific and cultural organizations. He is the Chairman of the Industrial Research Fund of the KU Leuven (\url{http://www.kuleuven.be/industrialresearchfund/}) and the Hercules foundation (heavy equipment funding in Flanders, \url{www.herculesstichting.be}). As a vice-rector for International Policy (2009--2013), he was a member of the Executive Committee and the Academic Council of the KU Leuven and of the Board of Directors of the Association KU Leuven. He also was the Chairman of the Committee on Internationalisation and Development Cooperation of the VLUHR (\url{www.vluhr.be}). He made regular television appearances in the Science Show ‘Hoe?Zo’ on national television in Belgium (\url{www.tv1.be}), has a regular science talk on Radio 2 (\url{www.radio2.be}) and patrons the ``Flemish Youth Technology Olympiade" (\url{http://www.technologieolympiade.be/vto/}). Full details on his CV can be found at \url{www.bartdemoor.be}.

\end{document}